\documentclass[lettersize,journal]{IEEEtran}
\usepackage{amsmath,amsfonts}
\usepackage{algorithmic}
\usepackage{algorithm}
\usepackage{array}
\usepackage{subfigure}
\usepackage[caption=false,font=normalsize,labelfont=sf,textfont=sf]{subfig}
\usepackage{textcomp}
\usepackage{stfloats}
\usepackage{url}
\usepackage{verbatim}
\usepackage{graphicx}
\usepackage{cite}
\newtheorem{assumption}{Assumption}

\begin{document}
\title{Adaptive Neural Network Backstepping Control Method for Aerial Manipulator Based on Variable Inertia Parameter Modeling}
\author{Hai Li, Zhan Li, \IEEEmembership{Member, IEEE}, Xiaolong Zheng, and Jinhui Liu
\thanks{This work was supported partially by the National Natural Science Foundation of China under Grant 62273122 and Grant U21B6001, and the Heilongjiang Natural Science Foundation under Grant LH2019F020. (Corresponding author: Zhan Li)}
\thanks{Hai Li, Zhan Li, Xiaolong Zheng and Jinhui Liu are with the Research Institute of Intelligent Control and Systems, Harbin Institute of Technology, Harbin 150001, China. Zhan Li is also with Peng Cheng Laboratory (e-mail: 18B904011@stu.hit.edu.cn; zhanli@hit.edu.cn; xiaolongzheng@hit.edu.cn; yzzxljh@126.com).}
}

%

\maketitle

\begin{abstract}
	For the aerial manipulator that performs aerial work tasks, the actual operating environment it faces is very complex, and it is affected by internal and external multi-source disturbances. In this paper, to effectively improve the anti-disturbance control performance of the aerial manipulator, an adaptive neural network backstepping control method based on variable inertia parameter modeling is proposed. Firstly, for the intense internal coupling disturbance, we analyze and model it from the perspective of the generation mechanism of the coupling disturbance, and derive the dynamics model of the aerial manipulator system and the coupling disturbance model based on the variable inertia parameters. Through the proposed coupling disturbance model, we can compensate the strong coupling disturbance in a way of feedforward. Then, the adaptive neural network is proposed and applid to estimate and compensate the additional disturbances, and the closed-loop controller is designed based on the backstepping control method. Finally, we verify the correctness of the proposed coupling disturbance model through physical experiment under a large range motion of the manipulator. Two sets of comparative simulation results also prove the accurate estimation of the proposed adaptive neural network for additional disturbances and the effectiveness and superiority of the proposed control method.
\end{abstract}


\section{Introduction}
\label{introduction}
\IEEEPARstart{W}{ith} the rapid development of unmanned aerial vehicle (UAV) in recent years, UAV has been widely utilized in aerial photography, surveying and mapping, search and rescue and other practical scenes because of its flexibility in three-dimensional space. However, most of these applications are simple collection of environmental information remotely without interacting with the environment. In order to further expand the application of UAV, in recent years, people hope to combine UAV and manipulator ingeniously to form a new type of robot system that has both flexible maneuverability in three-dimensional space and strong manipulation capabilities. In this context, the aerial manipulator system (AMS) appears and has attracted the attention and extensive research of more and more researchers and institutions in the world\cite{8299552,khamseh2018aerial}. AMS has a very broad application prospect and research value because of its superior characteristics. For example, various kinds of AMS equipped with different manipulators and end-effectors are designed and applied to valve turning\cite{6943037}, aerial maintenance\cite{8641488}, autonomous non-destructive contact inspection of industrial equipments\cite{8629273}, canopy sampling\cite{7152326}. In addition, the ARCAS and AEROARMS projects funded by the European Union have also conducted extensive research on the application scenarios of AMS in aerial assembly and structural construction\cite{baizid2017behavioral,muscio2017coordinated,lippiello2015hybrid}.

The strong coupling disturbance problem is one of the biggest challenges for AMS to accurately perform aerial operations, which is mainly caused by the change of the system center of mass and the moment of inertia due to the relative motion between the UAV and the manipulator, especially in the scenario where rapid and large range relative motion is required. The intense coupling disturbance will strongly affect the stability and the performance of AMS when performing aerial operation tasks and even let it out of control. In recent years, the research on the AMS focuses more attention on its application in various scenarios. Some early studies directly ignore or rely on the robustness of PID and other control algorithms \cite{chmaj2013dynamics,liu2021approximation,ghadiok2011autonomous} to deal with the coupling disturbance problem. Subsequently, some studies on coping with the coupling disturbance problem of AMS have also been published. In \cite{baizid2017behavioral,muscio2017coordinated,lippiello2015hybrid}, a mechanical device is designed to adjust the position of the battery tray to compensate for the change of the center of mass position of the system caused by the movement of the manipulator. In \cite{kondak2013closed} and \cite{buzzato2018aerial}, the authors directly measure the coupling disturbance through external force and torque sensors, and carries out feedforward compensation in the design of the controller. In \cite{kim2017robust,fanni2017new,chen2022adaptive}, the authors design controller based on disturbance observers, which make use of the system states to estimate the coupling disturbance force and torque and compensate them in the way of feedback. In \cite{huber2013first,jimenez2013control,heredia2014control,jimenez2016modelling}, considering the system centroid offset caused by the relative motion between UAV and manipulator, the aerial manipulator system dynamics is modeled, and a variable parameter integral backstepping controller is designed on this model to reject coupling disturbance. In \cite{liu2021ddpg}, the author proposed a method based on reinforcement learning DDPG algorithm to control the motion of the manipulator to ensure that it can bring about less coupling disturbance while tracking the desired trajectory.

Although the above works put forward a variety of solutions and methods, most of the research are carried out by designing specific mechanical mechanisms, resorting to additional torque sensor or restrictive planning for the relative motion of UAV and manipulator, rather than considering the generation mechanism of coupling disturbance. Through physical experiments, it is found that the above methods show some certain limitations when the relative motion between UAV and manipulator has large range and fast speed. Especially in some serious cases, when the manipulator moves rapidly in a large range, the severe coupling disturbance will seriously threaten the stability of the system, even let AMS be out of control. Therefore, based on variable inertia parameters \cite{zhang2019robust}, this paper deduces the dynamics model of the AMS, and derives a new coupling disturbance model, which can quantitatively describe the dominant part of the coupling disturbance of the AMS from the perspective of coupling disturbance generation mechanism. Based on the proposed coupling disturbance model, we can obtain the accurate coupling disturbance feedforward compensation to cope with the coupling disturbance without resorting to external force and torque sensors and related disturbance estimation methods.

For AMS performing aerial work tasks, the actual work environment it faces is very complex. In addition to the intense internal coupling disturbance, the AMS may also encounter additional disturbances caused by unmodeled dynamic terms and various uncertainties, such as sudden wind gust disturbance, additional gravitational disturbance after performing grasping tasks, etc., which also affect the performance of AMS for performing aerial work missions. The radial basis neural network (RBFNN) has the property of universal approximation, as well as the advantages of simple design, good generality, and strong online learning ability \cite{yu2011advantages,zhang2000adaptive}. Recently, learning-based adaptive control methods using neural networks have received increasing attentions in solving the disturbance rejection control problem of nonlinear systems with uncertainty \cite{cao2022adaptive,shi2022master,yang2018sgd}. Inspired by \cite{yang2019adaptive,tee2008adaptive}, by combining adaptive RBFNN and traditional feedback control methods, we propose and design a feedback compensation method based on adaptive neural network estimation, which can accurately estimate and compensate for these additional disturbances in real time and effectively improve the anti-disturbance control performance of the AMS. Finally, we propose an adaptive neural network backstepping control method based on variable inertia parameter modeling for AMS that encounters multi-source disturbances in the actual operating environment.

The rest of this paper is arranged as follows. In section \ref{Dynamics modeling of AMS}, the dynamic modeling of AMS based on variable inertia parameters is analyzed and described, and the coupling disturbance model based on variable inertia parameters is derived. Then, in section \ref{Adaptive neural network backstepping controller design}, the design process of adaptive neural network backstepping controller is given. Subsequently, the feasibility and effectiveness of the proposed method is verified by experiment and simulation, and the results and analysis are shown in section \ref{Experiment and simulation results}. The section \ref{Conclusion} concludes this paper and propose an outlook for future work. At the end of this section, the main contributions of this paper include the following aspects as follow:
\begin{enumerate}
	\item A new coupling disturbance model based on the variable inertia parameters of AMS is proposed, which makes full use of the state variables, state derivatives and second-order derivatives of UAV and manipulator, and also can fully reflect the static and dynamic relationship of coupling disturbance between UAV and manipulator.
	\item A complete verification physical AMS has been developed. Through the AMS platform, we verify the accuracy of the proposed coupled disturbance model under a large range motion of the manipulator. 
	\item In order to estimate and deal with the additional disturbances caused by unmodeled items and various uncertainties, a feedback compensation method based on adaptive neural network estimation is proposed. The simulation results show that the proposed method can quickly and accurately estimate other additional disturbances.
	\item An adaptive neural network backstepping control method based on variable inertia parameter modeling is proposed to effectively deal with the multi-source disturbances problem faced by the AMS in the actual aerial operating environment, which is based on coupling disturbance modeling for feedforward compensation and adaptive neural network estimation for feedback compensation.
\end{enumerate}

\section{Dynamics modeling of AMS}
\label{Dynamics modeling of AMS}
\subsection{Dynamic modeling of AMS}
When modeling the dynamics of the AMS, we no longer regard it as a single rigid body, but as a system composed of multiple rigid bodies. Assuming that the mass center of the UAV is at its geometric center, we can use the momentum theorem and momentum moment theorem of the multi-rigid body point system for system dynamics modeling.
\begin{figure}[htbp]
	\centering
	\includegraphics[width=0.45\textwidth]{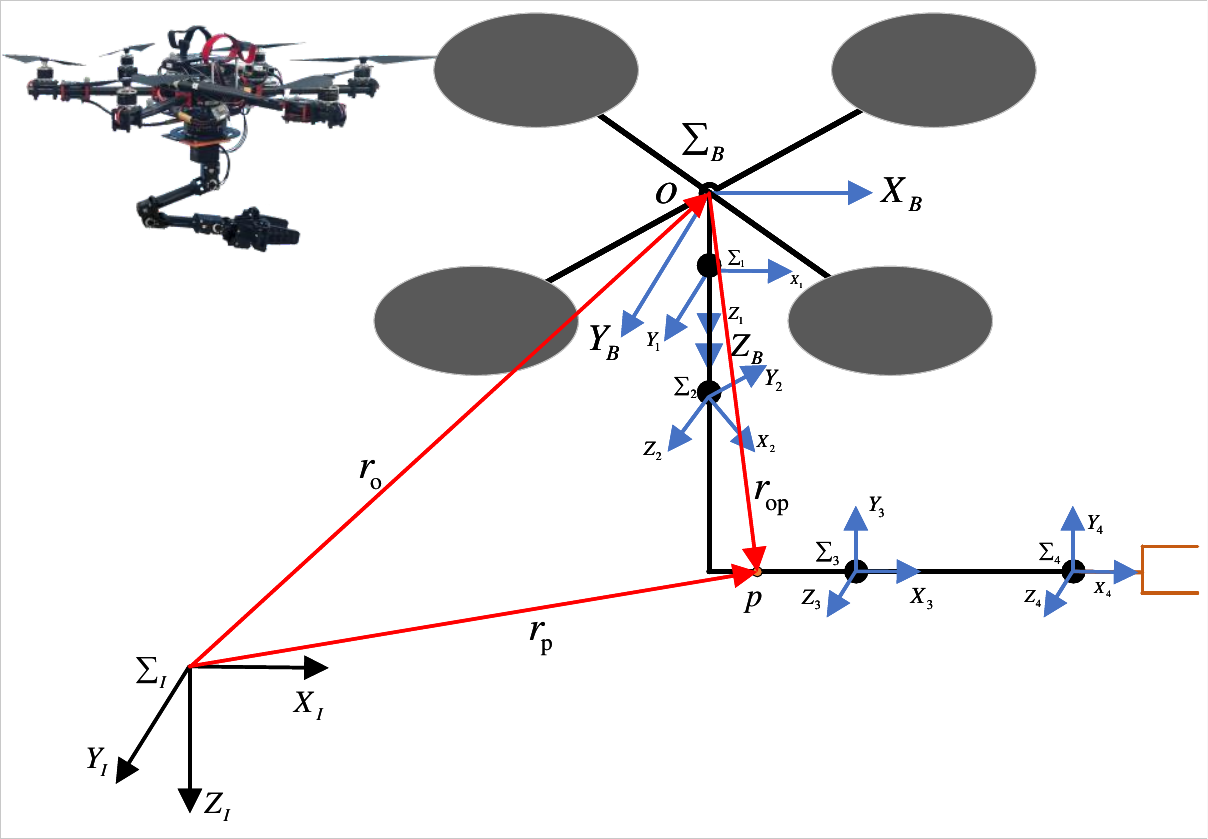}  
	\caption{The coordinate frame of AMS \label{coordinate_system}}
\end{figure}

In this paper, we consider that the AMS is composed of a quadrotor and a 4-DOF manipulator, and its coordinate system is established as shown in Fig. \ref{coordinate_system}, where $\Sigma_{I}$ and $\Sigma_{B}$ represent the inertial coordinate system (NED) and the body fixed coordinate system respectively ($X_{B}$-axis points to the UAV head direction, $Z_{B}$-axis points to the ground, and the original point $o$ of the coordinate system is located at the centroid of the UAV). $\Sigma_{i(i=1,2,3,4)}$ represents each link coordinate system of the manipulator, which are established based on the improved DH parameter method \cite{craig2005introduction}.
Assuming that point $p$ is any mass point in the AMS, we have
\begin{equation}
r_{p} = r_{o} + r_{op} = r_{o} + ^{I}R_{B}(^{B}r_{op})
\end{equation}
where, the vector $r_{p}$ represents absolute position of point $p$ relative to $\Sigma_{I}$, the vector $r_{o}$ represents absolute position of UAV relative to $\Sigma_{I}$, $r_{op}$ is the vector from point $o$ to point $p$, $^{B}r_{op}$ is the indication of $r_{op}$ with respect to $\Sigma_{B}$, $^{I}R_{B}$ represents the transformation matrix from $\Sigma_{B}$ to $\Sigma_{I}$,
\begin{equation}
^{I}R_{B}=
\begin{bmatrix}
{c\psi c\theta} & {-s \psi c \phi+c \psi s \theta s \psi} & {s \psi s \phi+c \psi s \theta c \phi} \\ 
{s \psi c \theta} & {c \psi c \phi+s \psi s \theta s \phi} & {-c \psi s \phi+s \psi s \theta c \phi} \\ 
{-s \theta} & {c \theta s \phi} & {c \theta c \phi}
\end{bmatrix}
\end{equation}
where, ${\Phi _b} = [\phi ,\theta ,\psi ]$, denoting roll, pitch, yaw angle respectively, are used to describe quadrotor attitude in the $Z-Y-X$ Euler angle. ${s,c}$ represent trigonometric function ${sin()}$ and ${cos()}$ respectively.

Then, we can obtain the momentum and moment of momentum of the AMS as follows,\\
\begin{equation}
\left\{
\begin{aligned}
P =& \int {_{{m_b} + {m_{man}}}} {{\dot r}_p}d{m_p} \\
=& {m_b}{{\dot r}_o} + {m_{man}}{{\dot r}_o} + {m_{man}}{}^I{R_B}\left( {{}^B{\omega _b} \times {}^B{r_{omc}} + {}^B{{\dot r}_{omc}}} \right) \\
L =& \int {_{{m_b} + {m_{man}}}} {r_p} \times {{\dot r}_p}d{m_p} \\
=& {r_o} \times P + {m_s}{r_{oc}} \times {{\dot r}_o} + {}^I{R_B}\left( {{}^B{I_b} + {}^BI_{man}^o} \right){}^B{\omega _b}\\
&+ {m_{man}}{}^I{R_B}({}^B{r_{omc}} \times {}^B{{\dot r}_{omc}})
\end{aligned}
\right.
\label{momentum}
\end{equation}
where, $m_{s},m_{b},m_{man}$ represent the total mass of the system, the mass of the UAV and the mass of the manipulator respectively. $P,L$, respectively, denote the momentum and moment of momentum of the system. $^{B}\omega_{b}$ stands for body angular velocity vector of UAV. $^{B}r_{omc}$ is the centroid vector of manipulator with respect to $\Sigma_{B}$. $r_{oc}$ is the centroid vector of the AMS with respect to $\Sigma_{I}$ and $^{B}r_{oc}$ is the representation of $r_{oc}$ in $\Sigma_{B}$. $^{B}I_{b}$ is the inertia matrix of UAV. $^{B}I^{o}_{man}$ represents the expression of the inertia matrix of the manipulator relative to point $o$ in $\Sigma_{B}$. $\times$ stands for cross product operation.

According to the theorem of momentum and the theorem of moment of momentum, we can have
\begin{equation}
\left\{
\begin{aligned}
\frac{{{d_P}}}{{{d_t}}} = &{}^I{F_{ext}}\\
\frac{{{d_L}}}{{{d_t}}} = &{r_o} \times {}^I{F_{ext}} + {}^IM_{ext}^o
\end{aligned}
\right.
\label{momentum_theory}
\end{equation}
where, $^{I}F_{ext}$ and $^{I}M^{o}_{ext}$ respectively represent the combined external force acting on the system and its resultant moment relative to point $o$ in the $\Sigma_{I}$. For the aerial manipulator system, they are
\begin{equation}
\left\{
\begin{aligned}
{}^I{F_{ext}} &=  - {F_l}{}^I{R_B}{e_3} + {m_s}g{e_3}\\
{}^IM_{ext}^o &= {}^I{R_B}\tau  + {m_s}{}^I{R_B}{}^B{r_{oc}} \times g{e_3}
\end{aligned}
\right.
\label{external_force_torque}
\end{equation}
where, $F_{l}$ and $\tau$ represent the lift and torque generated by the propellers of UAV respectively.

Taking (\ref{external_force_torque}) and (\ref{momentum}) into (\ref{momentum_theory}), the dynamics model of AMS can be obtained as follows
\begin{equation}
\left\{
\begin{aligned}
&{{\dot v}_b} = - \frac{{{F_t}}}{{{m_s}}}{}^I{R_B}{e_3} - \frac{{{m_{man}}}}{{{m_s}}}{}^I{R_B}({}^B{\omega _b} \times ({}^B{\omega _b} \times {}^B{r_{omc}})\\
&{\rm{       }} + {}^B{{\dot \omega }_b} \times {}^B{r_{omc}} + 2{}^B{\omega _b} \times {}^B{{\dot r}_{omc}} + {}^B{{\ddot r}_{omc}}) + g{e_3}\\
&({I_b} + {}^BI_{man}^o){}^B{{\dot \omega }_b} = \tau  - {}^B{\omega _b} \times (({I_b} + {}^BI_{man}^o){}^B{\omega _b}) + \\
& {m_s}({}^B{r_{oc}} \times ({}^B{R_I}g{e_3}) - {}^B{r_{oc}} \times {}^B{{\ddot r}_o} - {}^B{{\dot r}_{oc}} \times {}^B{{\dot r}_o}) - \\
&{m_{man}}({}^B{{\dot r}_o} \times ({}^B{\omega _b} \times {}^B{r_{omc}}) + {}^B{{\dot r}_o} \times {}^B{{\dot r}_{omc}} + \\
&{}^B{\omega _b} \times ({}^B{r_{omc}} \times {}^B{{\dot r}_{omc}}) + {}^B{r_{omc}} \times {}^B{{\ddot r}_{omc}}) - {}^B\dot I_{man}^o{}^B{\omega _b}
\end{aligned}
\right.
\label{dynamics_model}
\end{equation}
\subsection{Coupling disturbance model based on variable inertia parameters}
The motion of the manipulator lead to the change of the center of mass $^{B}r_{oc}$ and inertia matrix $^{B}I^{o}_{man}$ of the AMS, and the more intense the motion of the manipulator, the stronger the change of the center of mass and inertia of the system. The mapping relationship between this change and the coupling disturbance between the UAV and the manipulator is the coupling disturbance model we need to establish. The variable inertia parameters\cite{zhang2019robust} in the system are $^{B}r_{oc}$ and $^{B}I^{o}_{man}$, which are determined by the state variables of the manipulator, and their mapping relationship is as follows:
\begin{equation}
\left\{
\begin{aligned}
&{}^B{r_{oc}} = \frac{1}{{{m_s}}}\mathop \sum \limits_{i = 1}^n {m_i}^B{p_{ci}}\\
&{}^B{r_{omc}} = \frac{{{m_s}}}{{{m_{man}}}}{}^B{r_{oc}}\\
&{}^B{{\dot r}_{oc}} = \frac{1}{{{m_s}}}\mathop \sum \limits_{i = 1}^n {m_i}^B{v_{ci}}\\
&{}^B{p_{ci}} = {}^B{T_i}{(q)^i}{r_{ci}}\\
&{}^BI_{man}^o = \mathop \sum \limits_{i = 1}^4 ({}^B{R_i}I_i^{ci}{}^BR_i^{ - 1} + {m_i}(\parallel {}^B{p_{ci}}{\parallel ^2}{I_{3 \times 3}} -\\
&\ \ \ \ \ \ \ \ \ \ \ \ \ \ \ \ \ \ {}^B{p_{ci}}{({}^B{p_{ci}})^T}))\\
&{}^B\dot I_{man}^o = \mathop \sum \limits_{i = 1}^4 (Skew{(^B}{\omega _i}){}^B{R_i}I_i^{ci}{}^i{R_B} - {}^B{R_i}I_i^{ci}{}^i{R_B}Skew({}^B{\omega _i}))\\
&\ \ \ \ \ \ \ \ \ \ \ + \mathop \sum \limits_{i = 1}^4 {m_i}(2{({}^B{p_{ci}})^T}{}^B{v_{ci}}{I_{3 \times 3}} - {}^B{v_{ci}}{({}^B{p_{ci}})^T} -\\ 
&\ \ \ \ \ \ \ \ \ \ \ \ \ \ \ \ \ \ \ \ \ \ \ \ {}^B{p_{ci}}{({}^B{v_{ci}})^T})\\
&\left[ {\begin{array}{*{20}{c}}
	{{}^B{v_{ci}}}\\
	{{}^B{\omega _i}}
	\end{array}} \right] = {}^B{J_{ci}}(q)\dot q
\end{aligned}
\right.
\label{variable_inertia_parameters}
\end{equation}
where, ${m_{i(i = 1,2,3,4)}}$ are the mass of each link of manipulator. ${}^B{p_{ci}}$ and $^B{v_{ci}}$ denote the position and velocity of the centroid of link $i$ with respect to $\Sigma_{B}$ respectively. $q,\dot q$ respectively represent the angle and angular velocity of each joint of the manipulator. ${}^B{T_i}$ stands for the transformation matrix from $\Sigma_{i}$ to $\Sigma_{B}$, which can be obtained by the DH parameter. $^i{r_{ci}}$ represents the centroid of link $i$ in $\Sigma_{i}$. ${}^B{R_i}$ represents the transformation matrix from $\Sigma_{i}$ to $\Sigma_{B}$. $I_i^{ci}$ denotes the inertia matrix of the link $i$ in $\Sigma_{i}$. $^B{\omega _i}$ represents the angular velocity of the link $i$ with respect to $\Sigma_{B}$. $Skew{(^B}{\omega _i})$ is skew symmetric matrix of $^B{\omega _i}$. ${I_{3 \times 3}}$ is the identity matrix. ${}^B{J_{ci}}$ represents the jacobian matrix of link $i$ with respect to $\Sigma_{B}$.

By simplifying the system dynamics model of the above equation (\ref{dynamics_model}), the coupled disturbance model based on variable inertia parameters can be obtained as follows:
\begin{equation}
\left\{
\begin{aligned}
{F_{dis}} =  &- {m_{man}}{}^I{R_B}({}^B{\omega _b} \times ({}^B{\omega _b} \times {}^B{r_{omc}}) + \\
&{}^B{{\dot \omega }_b} \times {}^B{r_{omc}}{\rm{         }} + 2{}^B{\omega _b} \times {}^B{{\dot r}_{omc}} + {}^B{{\ddot r}_{omc}})\\
{}^B{\tau _{dis}} =& {m_s}({}^B{r_{oc}} \times ({}^B{R_I}g{e_3}) - {}^B{r_{oc}} \times {}^B{{\ddot r}_o} - {}^B{{\dot r}_{oc}} \times {}^B{{\dot r}_o}) \\
&- {m_{man}}({}^B{{\dot r}_o} \times {}^B{{\dot r}_{omc}}{\rm{          }} + {}^B{{\dot r}_o} \times ({}^B{\omega _b} \times {}^B{r_{omc}}) + \\
&{}^B{\omega _b} \times ({}^B{r_{omc}} \times {}^B{{\dot r}_{omc}}) + {}^B{r_{omc}} \times {}^B{{\ddot r}_{omc}}) - \\
&{}^B\dot I_{man}^o{}^B{\omega _b} - {}^B{\omega _b} \times ({}^BI_{man}^o{}^B{\omega _b}) - {}^BI_{man}^o{}^B{{\dot \omega }_b}
\end{aligned}
\right.
\label{coupling_disturbance_model}
\end{equation}

It is not hard to see from the above (\ref{coupling_disturbance_model}) that the coupling disturbance model based on variable inertia parameters includes the state variables, state variable derivatives and their second derivative of UAV and manipulator. Theoretically, it makes full use of the state variables of the system, and these state variables can be measured directly through the sensor or estimated indirectly, which can well reflect the static and dynamic relationship of coupling disturbance between UAV and manipulator.

\section{Adaptive neural network backstepping controller design}
\label{Adaptive neural network backstepping controller design}
\begin{figure}[htbp]
	\centering
	\includegraphics[width=0.48\textwidth]{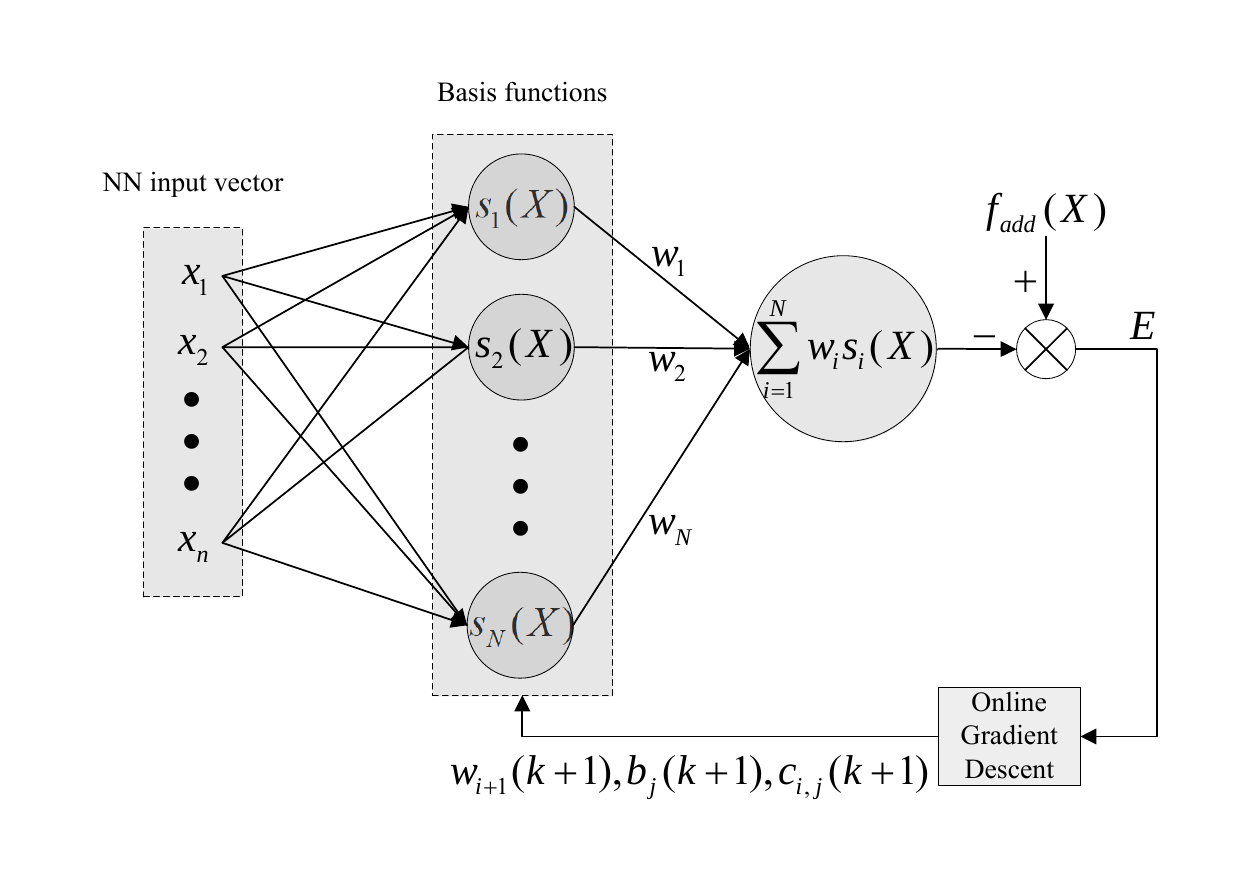}  
	\caption{The structure of the RBFNN with OGD algorithm \label{RBFNN_with_OGD_structure}}
\end{figure}

In this section, the design process of the adaptive neural network backstepping controller for aerial manipulator is described in detail. In the process of controller design, we introduced radial basis function neural network (RBFNN) and used online gradient descent (OGD) algorithm \cite{biehl1995learning} for online training, which can estimate and compensate the additional disturbances caused by unmodeled dynamics items and uncertainties in real time online. The structure of the RBFNN with OGC algorithm is given in Fig. \ref{RBFNN_with_OGD_structure}.

In Fig. \ref{RBFNN_with_OGD_structure}, $X=[x_{1},x_{2},\cdot\cdot\cdot,x_{n}]^\mathrm{T}$ represents the NN input vector. $S(X) = {[{s_1}(X),{s_2}(X), \cdots ,{s_N}(X)]^T}$ is the basis function vector of RBFNN and $N$ denotes the node number of a neural network. $W=[w_{1},w_{2},\cdot\cdot\cdot,w_{N}]^\mathrm{T}$ stands for the weight vector of RBFNN. ${f_{add}}(X)$ represents the additional disturbances to be estimated. $E = {f_{add}}(X) - {W^T}S(X)$ denotes the NN approximation error. In this paper, we choose the frequently used Gaussian Kernel function as the basis function ${s_i}(X)$
\begin{equation}
	{s_i}(X) = \exp \left( { - \frac{{{{(X - {C_i})}^T}(X - {C_i})}}{{b_i^2}}} \right)
\end{equation}
where ${b_j}$ and ${C_i} = {[{c_{1i}},{c_{2i}}, \cdots ,{c_{ni}}]^T}$ represent the width and center vectors of the Gaussian basis function, respectively.

\begin{assumption}
	In order to facilitate the subsequent design of attitude loop controller, we have made the assumption that the conversion relationship between body angular velocity and attitude angular velocity is ignored, that is, the difference between body angular velocity and attitude angular velocity is ignored. The assumption is reasonable when the roll angle and pitch angle change in a small angle range.
\end{assumption}
\begin{assumption}
	It is assumed that the UAV is axisymmetric, that is, there is no inertia product term in the moment of inertia matrix.
\end{assumption}

According to the above derivation results, let $x_{1} = x, x_{2} = \dot{x}, x_{3} = y, x_{4} = \dot{y}, x_{5} = z, x_{6} = \dot{z}, x_{7} = \phi, x_{8} = p, x_{9} = \theta, x_{10} = q, x_{11} = \psi, x_{12} = r$, the state-space model of the AMS is in the form in (\ref{dynamics_model_in_state_space})
\begin{equation}
\left\{
\begin{aligned}
{{\dot x}_1} &= {x_2}\\
{{\dot x}_2} &= \frac{1}{m}({u_x} + {{\hat F}_{disx}}) + {f_{addx}}\\
{{\dot x}_3} &= {x_4}\\
{{\dot x}_4} &= \frac{1}{m}({u_y} + {{\hat F}_{disy}}) + {f_{addy}}\\
{{\dot x}_5} &= {x_6}\\
{{\dot x}_6} &= \frac{1}{m}({u_z} + {{\hat F}_{disz}}) + g + {f_{addz}}\\
{{\dot x}_7} &= {x_8}\\
{{\dot x}_8} &= \frac{1}{{{J_\phi }}}({\tau _\phi } + {}^B{{\hat \tau }_{dis\phi }} + {J_\theta }{x_{10}}{x_{12}} - {J_\psi }{x_{12}}{x_8}) + {f_{add\phi }}\\
{{\dot x}_9} &= {x_{10}}\\
{{\dot x}_{10}} &= \frac{1}{{{J_\theta }}}({\tau _\theta } + {}^B{{\hat \tau }_{dis\theta }} + {J_\psi }{x_{12}}{x_{10}} - {J_\phi }{x_8}{x_{12}}) + {f_{add\theta }}\\
{{\dot x}_{11}} &= {x_{12}}\\
{{\dot x}_{12}} &= \frac{1}{{{J_\psi }}}({\tau _\psi } + {}^B{{\hat \tau }_{dis\psi }} + {J_\phi }x_8^2 - {J_\theta }x_{10}^2) + {f_{add\psi }}
\end{aligned}
\right.
\label{dynamics_model_in_state_space}
\end{equation}
where $x=[x_{1},x_{2},\cdot\cdot\cdot,x_{12}]^\mathrm{T}$ are the system states vector. $[x,y,z]$ and $[\dot{x},\dot{y},\dot{z}]$ represent the position and velocity of UAV, $[\phi,\theta,\psi]$ and $[p,q,r]$ denote the attitude and attitude angular velocity of UAV. $\hat{F}_{dis}$ and $^{B}\hat{\tau}_{dis}$ stand for the coupling disturbance force and moment respectively. $[J_{\phi},J_{\theta},J_{\psi}]$ are the moment of inertia of UAV. $[u_{x},u_{y},u_{z}]$ and $[\tau_{\phi},\tau_{\theta},\tau_{\psi}]$ represent control force and control torque respectively. $[f_{addx}, f_{addy}, f_{addz}]$ and $[f_{add\phi}, f_{add\theta}, f_{add\psi}]$ represent the additional disturbance forces and moments caused by unmodeled dynamics and uncertainties respectively.
\subsection{The design of position loop controller}
Let $z_{1} = x_{1} - x_{d}, z_{2} = x_{2} - \alpha_{1}, z_{3} = x_{3} - y_{d}, z_{4} = x_{4} - \alpha_{2}, z_{5} = x_{5} - z_{d}, z_{6} = x_{6} - \alpha_{3}, \alpha_{1} = -k_{1}z_{1} + \dot{x}_{d}, \alpha_{2} = -k_{3}z_{3} + \dot{y}_{d}, \alpha_{3} = -k_{5}z_{5} + \dot{z}_{d}$ with $k_{i,i=1,3,5} > 0$. The Lyapunov function candidate is chosen as $V_{1} = \sum^{6}_{i=1} z^{2}_{i}/2$ whose time derivative can be obtained by
\begin{equation}
\begin{aligned}
{{\dot V}_1} = &{z_1}({z_2} + {\alpha _1} - {{\dot x}_d}) + {z_2}(\frac{1}{m}({u_x} + {{\hat F}_{disx}}) + {f_{addx}} - {{\dot \alpha }_1})\\
&+ {z_3}({z_4} + {\alpha _2} - {{\dot y}_d}) + {z_4}(\frac{1}{m}({u_y} + {{\hat F}_{disy}}) + {f_{addy}} - {{\dot \alpha }_2})\\
&+ {z_5}({z_6} + {\alpha _3} - {{\dot z}_d}) + {z_6}(\frac{1}{m}({u_z} + {{\hat F}_{disz}}) + {f_{addz}} - {{\dot \alpha }_3})\\
\end{aligned}
\label{V1_dot}
\end{equation}

Design the position controller as follow
\begin{equation}
\left\{
\begin{aligned}
{u_x} &= m( - {k_2}{z_2} + {{\dot \alpha }_1} - W_1^T{S_1}({X_1}) - {z_1}) - {{\hat F}_{disx}}\\
{u_y} &= m( - {k_4}{z_4} + {{\dot \alpha }_2} - W_2^T{S_2}({X_2}) - {z_3}) - {{\hat F}_{disy}}\\
{u_z} &= m( - {k_6}{z_6} + {{\dot \alpha }_3} - g - W_3^T{S_3}({X_3}) - {z_5}) - {{\hat F}_{disz}}
\end{aligned}
\right.
\label{position_controller}
\end{equation}
where $k_{1} \sim k_{6}$ are the controller parameters and are positive numbers. $W^{T}_{j},S_{j}(x_{j})(j=1,2,3)$ represent the radial basis function neural network weight and kernel function of each channel of the position loop respectively. $X_1,X_2,X_3$ represent the NN input vector. In this paper, we choose the system states vector $x$ as the input vector of $NN_{j}(j=1,2,3)$.

The conversion relationship between position and attitude is as follows
\begin{equation}
\left\{
\begin{aligned}
{u_x} &=  - {u_m}(\cos \phi \sin \theta \cos \psi  + sin\phi \sin \psi )\\
{u_y} &=  - {u_m}(\cos \phi \sin \theta \sin \psi  - sin\phi \cos \psi )\\
{u_z} &=  - {u_m}\cos \phi \cos \theta 
\end{aligned}
\right.
\label{pos_att_transform}
\end{equation}
According to the conversion relationship in (\ref{pos_att_transform}), we can calculate the desired thrust and desired pitch and roll angles as follows
\begin{equation}
\left\{
\begin{aligned}
{u_m} &= \sqrt {u_x^2 + u_y^2 + u_z^2} \\
{\phi _d} &= \arcsin [u_m^{ - 1}({u_y}\cos \psi  - {u_x}\sin \psi )]\\
{\theta _d} &= \arctan [u_z^{ - 1}({u_x}\cos \psi  + {u_y}\sin \psi )]
\end{aligned}
\right.
\label{desired_thrust_pitch_roll}
\end{equation}
Then, by taking the designed position controllers into $\dot{z}_{2}, \dot{z}_{4}, \dot{z}_{6}$ respectively, we can get the NN approximation error:
\begin{equation}
{f_i} - W_i^T{S_i}(X) = {\dot z_{2i}} + {k_{2i}}{z_{2i}} + {z_{2i - 1}}(i = 1,2,3)
\label{NN_output_position}
\end{equation}
According to the online gradient descent algorithm\cite{biehl1995learning}, we can get the update rate of the neural network in the position loop
\begin{equation}
\left\{
\begin{aligned}
{{\dot W}_i} &= {\eta _i}{E_i}{S_i}(X)\\
{E_i} &= {{\dot z}_{2i}} + {z_{2i - 1}} + {k_{2i}}{z_{2i}}(i = 1,2,3)
\end{aligned}
\right.
\label{NN_update_rate_position}
\end{equation}
where $\eta_{i} \in (0,1)$ stands for the learning rate, $E_{i}$ denotes the NN approximation error.
\subsection{The design of attitude loop controller}
Let $z_{7} = x_{7} - \phi_{d}, z_{8} = x_{8} - \alpha_{4}, z_{9} = x_{9} - \theta_{d}, z_{10} = x_{10} - \alpha_{5}, z_{11} = x_{11} - \psi_{d}, z_{12} = x_{12} - \alpha_{6}, \alpha_{4} = -k_{7}z_{7} + \dot{\phi}_{d}, \alpha_{5} = -k_{9}z_{9} + \dot{\theta}_{d}, \alpha_{6} = -k_{11}z_{11} + \dot{\psi}_{d}$ with $k_{i,i=7,9,11} > 0$. The Lyapunov function candidate is chosen as $V_{2} = \sum^{12}_{i=7} z^{2}_{i}/2$ whose time derivative can be obtained by
\begin{equation}
\begin{aligned}
{{\dot V}_2} = &{z_7}({z_8} + {\alpha _4} - {{\dot \phi }_d}) + {z_8}(\frac{1}{{{J_\phi }}}({\tau _\phi } + {}^B{{\hat \tau }_{dis\phi }} + {J_\theta }{x_{10}}{x_{12}}\\
&-{J_\psi }{x_{12}}{x_8}) + {f_{add\phi }} - {{\dot \alpha }_4}) + {z_9}({z_{10}} + {\alpha _5} - {{\dot \theta }_d}) + \\
&{z_{10}}(\frac{1}{{{J_\theta }}}({\tau _\theta } + {}^B{{\hat \tau }_{dis\theta }} + {J_\psi }{x_{12}}{x_{10}} - {J_\phi }{x_8}{x_{12}}) + {f_{add\theta }}\\
&- {{\dot \alpha }_5}) + {z_{11}}({z_{12}} + {\alpha _6} - {{\dot \psi }_d}) + {z_{12}}(\frac{1}{{{J_\psi }}}({\tau _\psi } + {}^B{{\hat \tau }_{dis\psi }}\\
&+ {J_\phi }x_8^2 - {J_\theta }x_{10}^2) + {f_{add\psi }} - {{\dot \alpha }_6})
\end{aligned}
\label{V2_dot}
\end{equation}

Design the attitude controller as
\begin{equation}
\left\{
\begin{aligned}
{\tau _\phi } =& {J_\phi }( - {k_8}{z_8} - W_4^T{S_4}({X_4}) - {z_7}) - {}^B{{\hat \tau }_{dis\phi }} \\
&- {J_\theta }{x_{11}}{x_{12}} + {J_\psi }{x_{12}}{x_8}\\
{\tau _\theta } =& {J_\theta }( - {k_{10}}{z_{10}} - W_5^T{S_5}({X_5}) - {z_9}) - {}^B{{\hat \tau }_{dis\theta }}\\
&- {J_\psi }{x_{12}}{x_{10}} + {J_\phi }{x_8}{x_{12}}\\
{\tau _\psi } =& {J_\psi }( - {k_{12}}{z_{12}} - W_6^T{S_6}({X_6}) - {z_{11}}) - {}^B{{\hat \tau }_{dis\psi }}\\
&- {J_\phi }x_8^2 + {J_\theta }x_{10}^2
\end{aligned}
\right.
\label{attitude_controller}
\end{equation}
where $k_{7} \sim k_{12}$ are the controller parameters and are positive numbers. $W^{T}_{j},S_{j}(x_{j})(j=4,5,6)$ represent the radial basis function neural network weight and kernel function of each channel of the attitude loop respectively. $X_4,X_5,X_6$ represent the NN input vector. In this paper, we choose the system states vector $x$ as the input vector of $NN_{j}(j=4,5,6)$.

Subsequently, by taking the designed attitude controllers into $\dot{z}_{8}, \dot{z}_{10}, \dot{z}_{12}$ respectively, we have
\begin{equation}
{f_i} - {\dot \alpha _i} - W_i^T{S_i}(X) = {\dot z_{2i}} + {k_{2i}}{z_{2i}} + {z_{2i - 1}}(i = 4,5,6)
\end{equation}
Then, we can get the update rate of the neural network in the attitude loop
\begin{equation}
\left\{
\begin{aligned}
{W_i} &= {\eta _i}{E_i}{S_i}(X)\\
{E_i} &= {{\dot z}_{2i}} + {z_{2i - 1}} + {k_{2i}}{z_{2i}}(i = 4,5,6)
\end{aligned}
\right.
\label{NN_update_rate_attitude}
\end{equation}
From (\ref{NN_update_rate_position}) and (\ref{NN_update_rate_attitude}), it can be seen that the output of neural network in the position loop is the estimated value of the additional disturbance, while the output of neural network in the attitude loop is the estimated value of the difference between the additional disturbance and the intermediate varible $\alpha_{i}$. This is because the derivative of the intermediate quantity cannot be obtained directly, so we deal with it through the neural network. 

\begin{figure*}[htbp]
	\centering
	\includegraphics[width=0.80\textwidth]{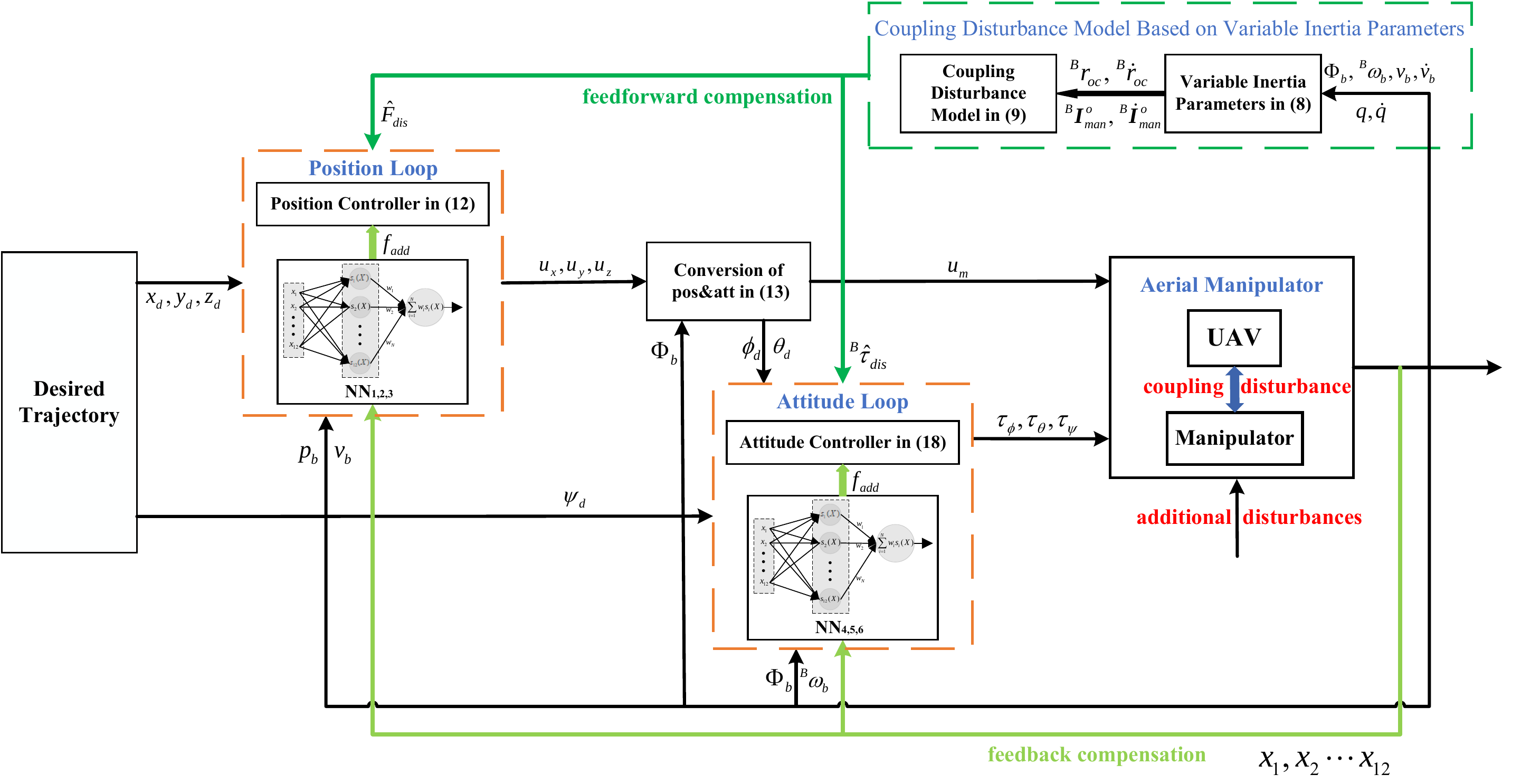}  
	\caption{The control scheme structure of aerial manipulator \label{control_structurer}}
\end{figure*}

The design of the adaptive neural network backstepping controller based on the variable inertia parameter modeling for the AMS has been completed, and the entire control block diagram can be seen in Fig. \ref{control_structurer}.

\section{Experiment and simulation results}
\label{Experiment and simulation results}
In order to verify the correctness and accuracy of the coupled disturbance model based on variable inertia parameters, experiment is carried out on the built physical platform. Then, for further verifying the effect of the control strategy proposed in this paper aimed at the coupled disturbance problem, we compared it with the PID control algorithm in the simulation.
\subsection{Experimental verification of coupled disturbance model}
\subsubsection{Experiment platform and conditions}
\begin{figure}[htbp]
	\centering
	\includegraphics[width=0.45\textwidth]{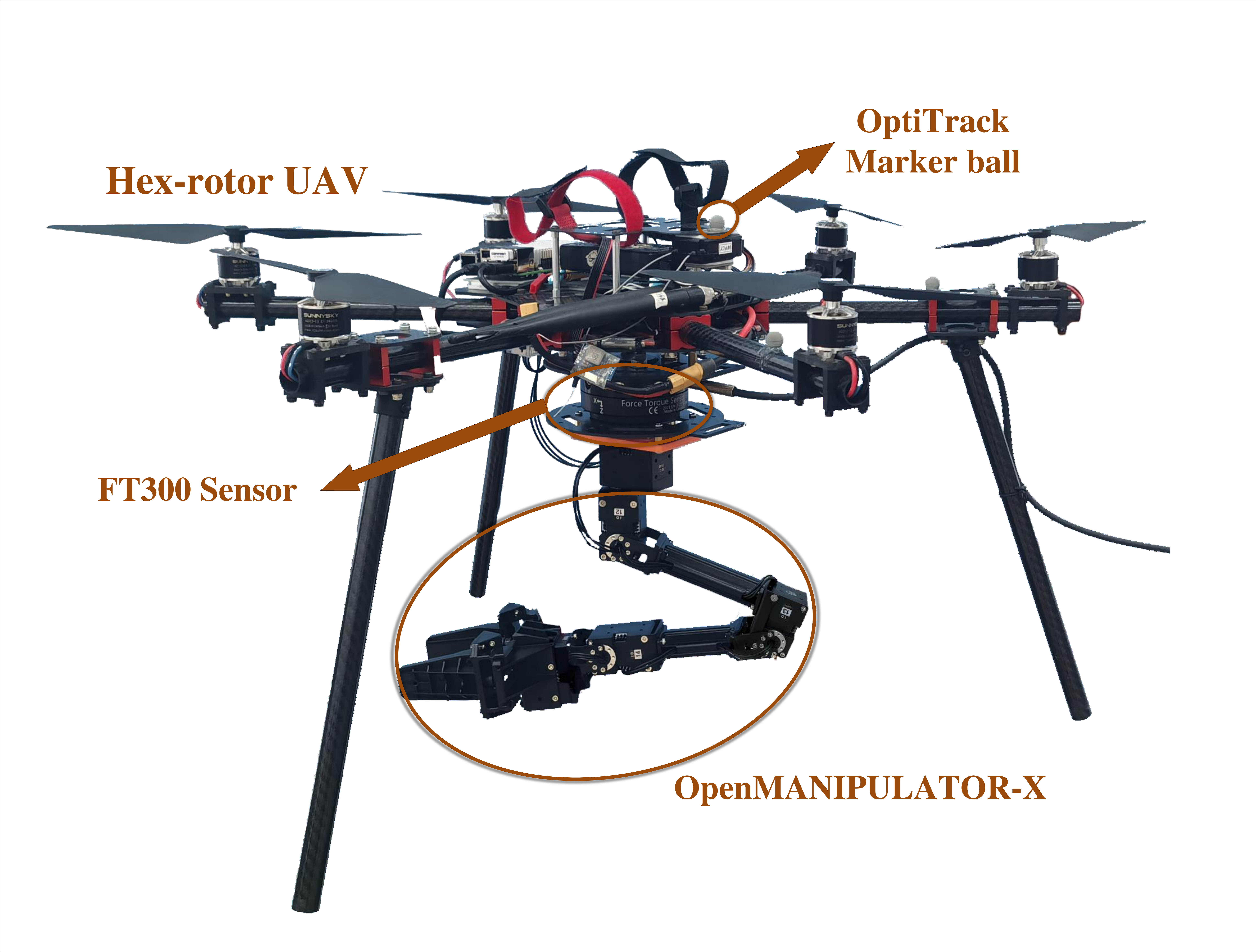}  
	\caption{Composition of aerial manipulator platform \label{experiment_platform}}
\end{figure}
In the experiment, the hex-rotor aerial manipulator platform we built and used is shown in Fig. \ref{experiment_platform}, which is mainly composed of a hex-rotor UAV, a 4-dof manipulator and a six-axis force and torque sensor. The physical parameters of the hex-rotor UAV are given as follow: $m_{b} = 2.65$, $J_\phi = 0.05$, $J_\theta = 0.05$, $J_\psi = 0.0948$, $l(wheelbase) = 0.55$. The manipulator is the open-source 4-dof manipulator (OpenMANIPULATOR-X) produced by ROBOTIS company and consists of five DYNAMIXEL XM430-W350-T actuators, which can provide high-precision joint angles, speeds and torque states information in real time. At the same time, the physical parameters of the 4-DOF manipulator are presented in Table \ref{physical_parameter_of_manipulator}. In addition, the kinematic modeling based on the improved DH parameters are shown in Table \ref{DH_parameter}. The six-axis force and torque sensor is the FT300 sensor produced by ROBOTIQ company, which is installed between the drone and the manipulator, and can directly and accurately measure the coupling disturbance force and torque between the drone and the manipulator in real time. Finally, the experiment was carried out under the indoor high-precision motion capture system (OptiTrack system), which can provide millimeter-level position, velocity and orientation information for the AMS.\\
\begin{table}[htb!]
	\caption{Physical Parameters of the 4-DOF manipulator}
	\label{physical_parameter_of_manipulator}
	\centering
	\begin{tabular}{|c|c|}
		\hline\rule{0pt}{12pt}
		$m_{mani}$ & 0.702kg \\
		\hline\rule{0pt}{12pt}
		$m_{1}$ & 0.238kg \\
		\hline\rule{0pt}{12pt}
		$m_{2}$ & 0.123kg \\
		\hline\rule{0pt}{12pt}
		$m_{3}$ & 0.118kg \\
		\hline\rule{0pt}{12pt}
		$m_{4}$ & 0.224kg \\
		\hline\rule{0pt}{12pt}
		$^1{r_{c1}}$ & (-0.006794,0.000253,-0.048813) \\
		\hline\rule{0pt}{12pt}
		$^2{r_{c2}}$ & (0.107084,-0.010616,0.000467) \\
		\hline\rule{0pt}{12pt}
		$^3{r_{c3}}$ & (0.094329,0.0000,0.000489) \\
		\hline\rule{0pt}{12pt}
		$^4{r_{c4}}$ & (0.060527,-0.006058,-0.000021) \\
		\hline\rule{0pt}{12pt}
		$I_1^{c1}$ & $10^{-4} * \left[ {\begin{array}{ccc}
			2.90202 & 0.00335 & 0.32543 \\
			0.00335 & 3.24158 & 0.02059 \\
			0.32543 & 0.02059 & 1.41275 \\
			\end{array} } \right]$ \\
		\hline\rule{0pt}{12pt}
		$I_2^{c2}$ & $10^{-4} * \left[ {\begin{array}{ccc}
			0.33028 & -0.06189 & 0.01212 \\
			-0.06189 & 1.84812 & -0.0002 \\
			0.01212 & -0.0002 & 1.89169 \\
			\end{array} } \right]$ \\
		\hline\rule{0pt}{12pt}
		$I_3^{c3}$ & $10^{-4} * \left[ {\begin{array}{ccc}
			0.20796 & 0.00002 & 0.01064 \\
			0.00002 & 1.45545 & 0.00 \\
			0.01064 & 0.00 & 1.38574 \\
			\end{array} } \right]$ \\
		\hline\rule{0pt}{12pt}
		$I_4^{c4}$ & $10^{-4} * \left[ {\begin{array}{ccc}
			1.43765 & 0.21123 & 0.00001 \\
			0.21123 & 2.12697 & 0.00485 \\
			0.00001 & 0.00485 & 1.80588 \\
			\end{array} } \right]$ \\
		\hline
	\end{tabular}
\end{table}
\begin{table}[htb!]
	\caption{Improved DH Parameters of the 4-DOF manipulator}
	\label{DH_parameter}
	\centering
	\begin{tabular}{ccccc}
		\hline
		\rule{0pt}{12pt} 
		$i$      & $\alpha_{i-1}$   & $a_{i-1}$   & $d_{i}$   & $\theta_{i}$\\[2pt]
		\hline\rule{0pt}{12pt}
		1     & 0    & 0.012    & 0.0935  & $\theta_{1}$\\
		\rule{0pt}{0pt}
		2     & $-\frac{\pi}{2}$  & 0   & 0    & $\theta_{2} - 1.3855$\\
		\rule{0pt}{0pt}
		3     & 0    & 0.13023    & 0   & $\theta_{3} + 1.3855$\\ 
		\rule{0pt}{0pt}
		4     & 0    & 0.124    & 0     & $\theta_{4}$\\[2pt]
		\hline
	\end{tabular}
\end{table}

\begin{figure}[htbp]
	\centering
	\subfigure[]{\includegraphics[width=0.23\textwidth]{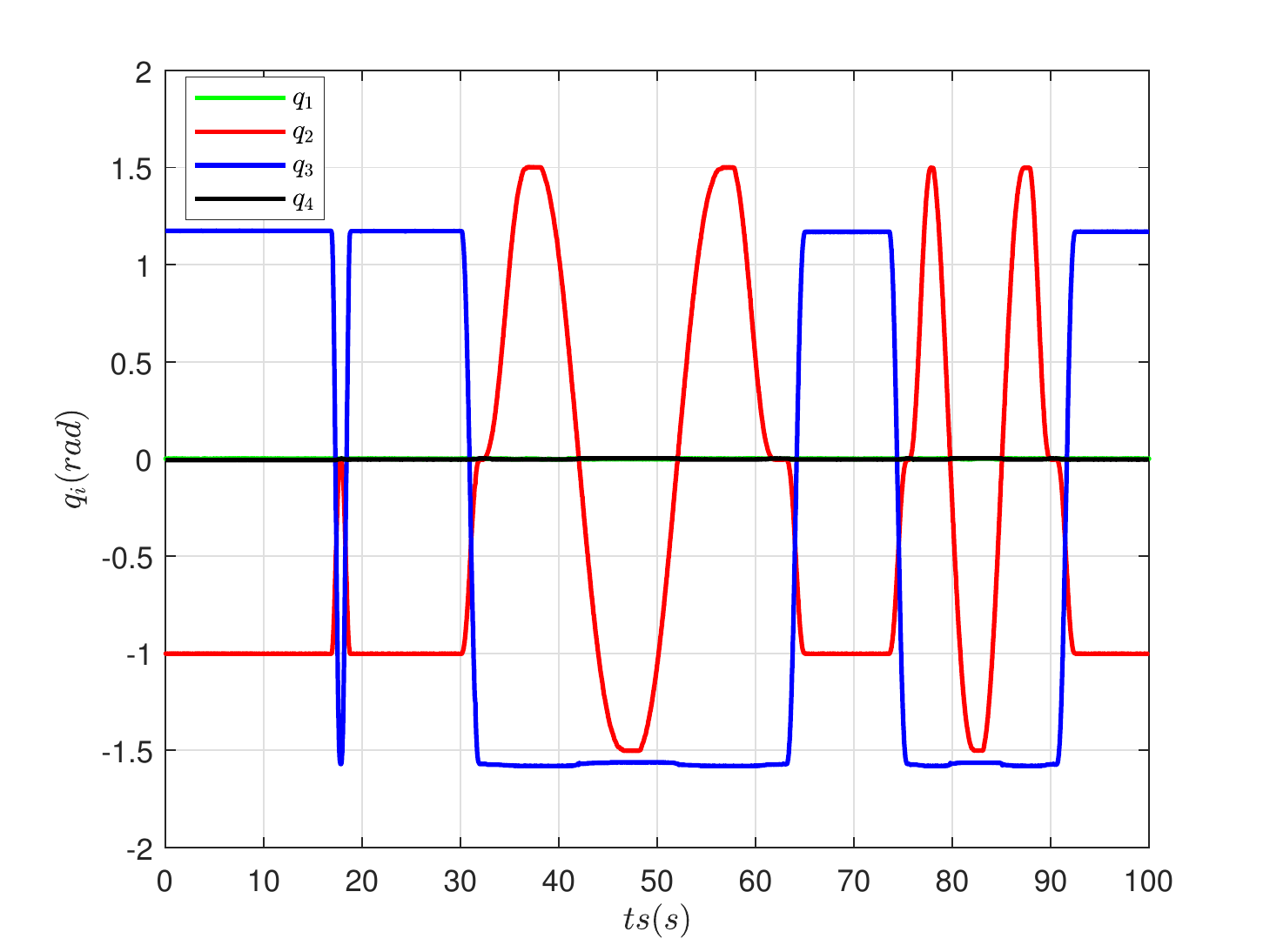}}
	\label{q_mani}
	\subfigure[]{\includegraphics[width=0.23\textwidth]{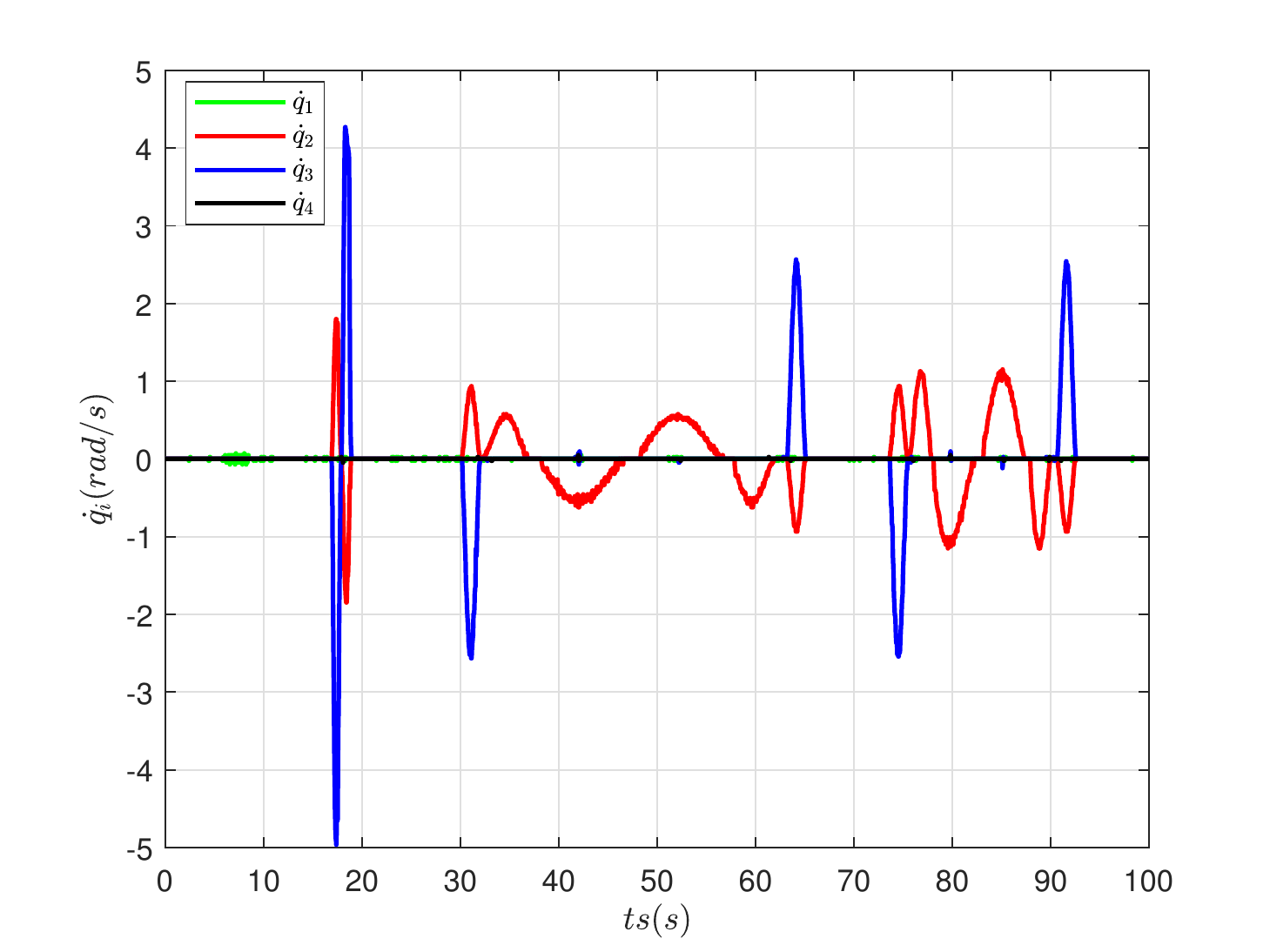}}
	\label{q_mani_dot}
	\caption{The joints trajectories and angular velocities in experiment \label{manipulator_motions}}
\end{figure}

\begin{figure*}[b]
	\begin{minipage}{0.16\linewidth}
		\vspace{2pt}
		\centerline{\includegraphics[width=1.0\textwidth]{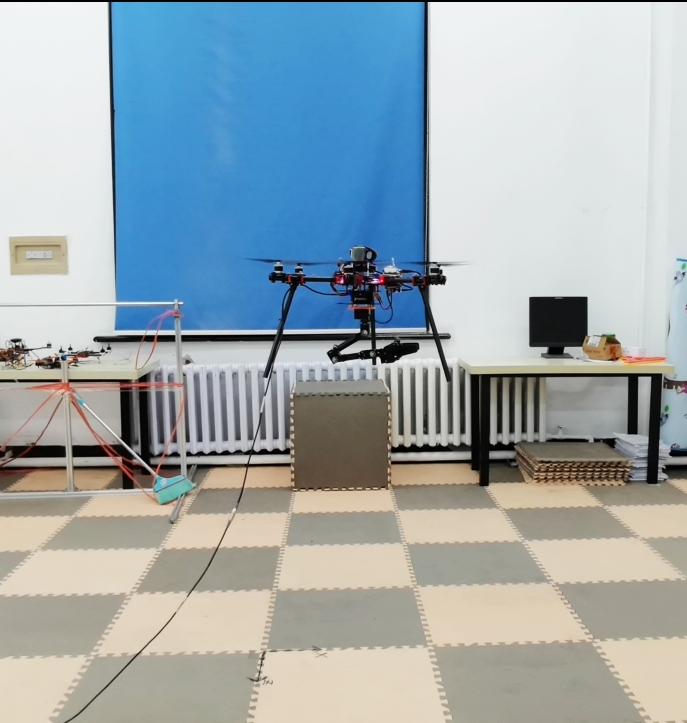}}
		\vspace{2pt}
		\centerline{\includegraphics[width=1.0\textwidth]{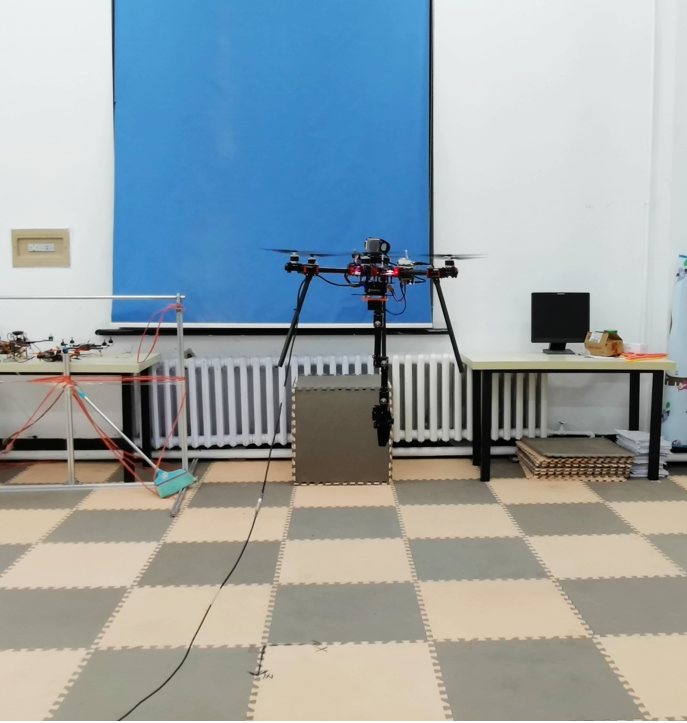}}
	\end{minipage}
	\begin{minipage}{0.16\linewidth}
		\vspace{2pt}
		\centerline{\includegraphics[width=1.0\textwidth]{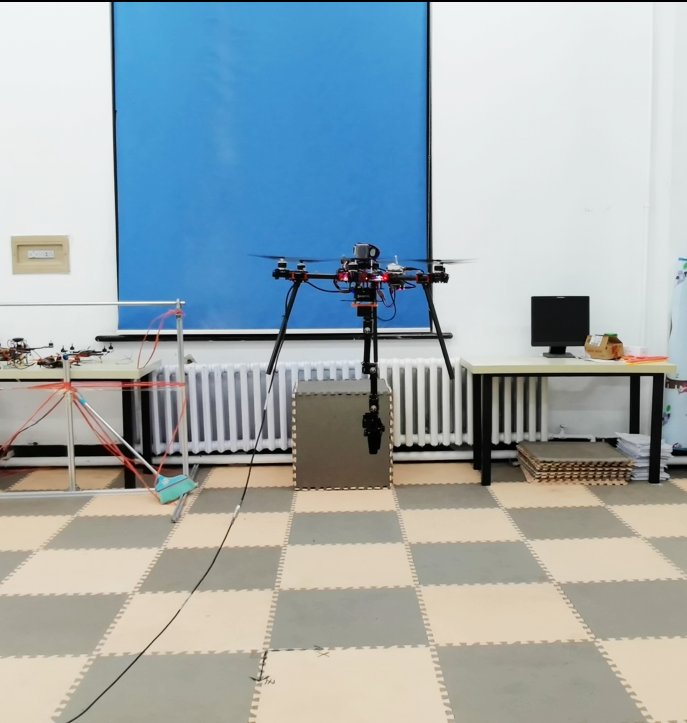}}
		\vspace{2pt}
		\centerline{\includegraphics[width=1.0\textwidth]{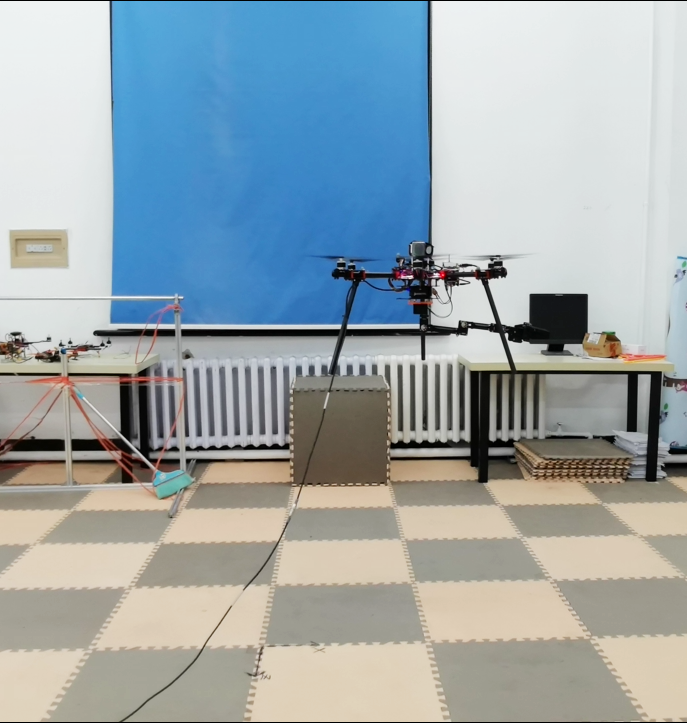}}
	\end{minipage}
	\begin{minipage}{0.16\linewidth}
		\vspace{2pt}
		\centerline{\includegraphics[width=1.0\textwidth]{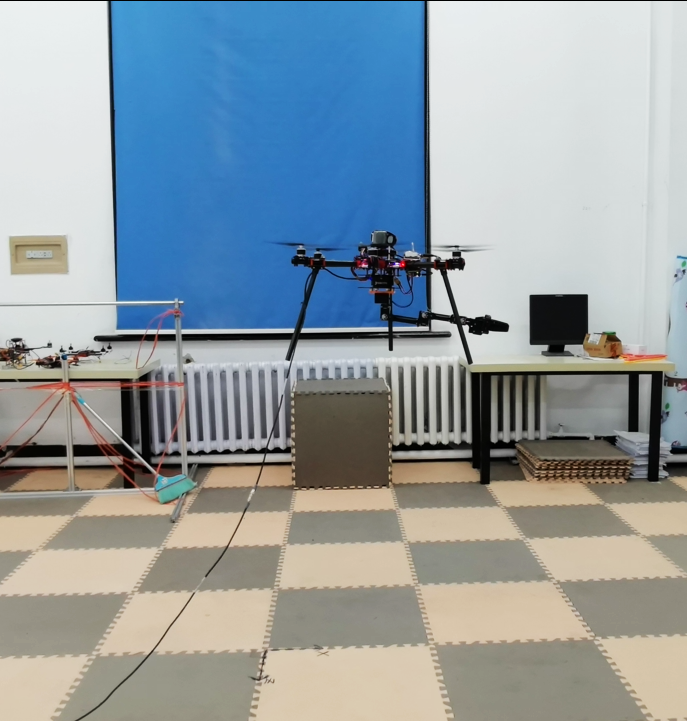}}
		\vspace{2pt}
		\centerline{\includegraphics[width=1.0\textwidth]{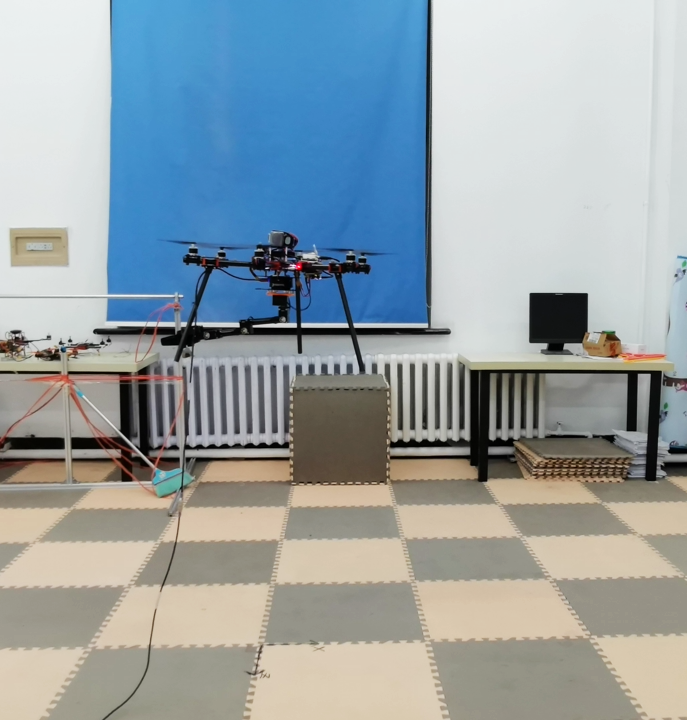}}
	\end{minipage}
	\begin{minipage}{0.16\linewidth}
		\vspace{2pt}
		\centerline{\includegraphics[width=1.0\textwidth]{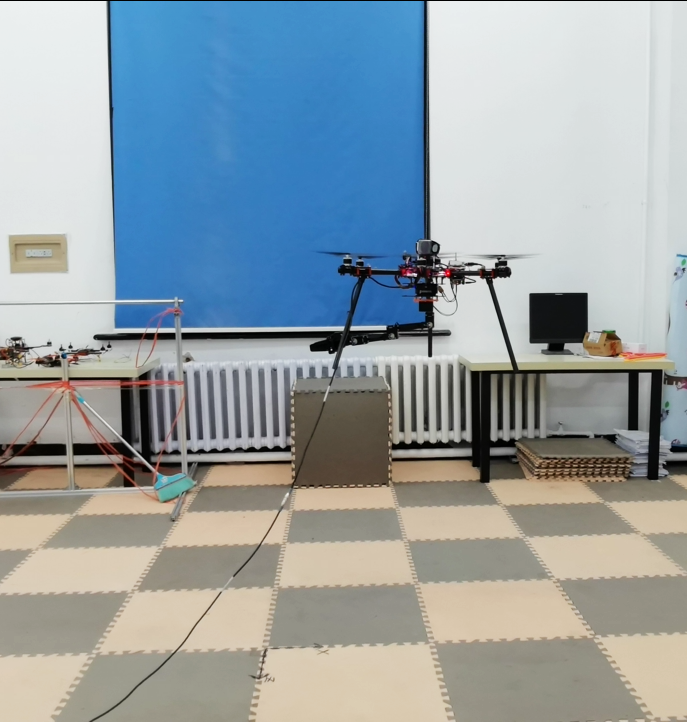}}
		\vspace{2pt}
		\centerline{\includegraphics[width=1.0\textwidth]{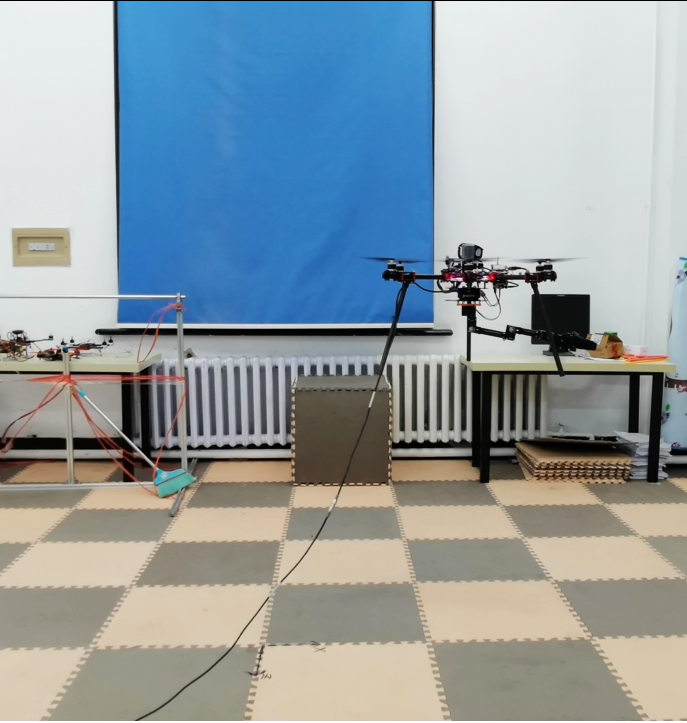}}
	\end{minipage}
	\begin{minipage}{0.16\linewidth}
		\vspace{2pt}
		\centerline{\includegraphics[width=1.0\textwidth]{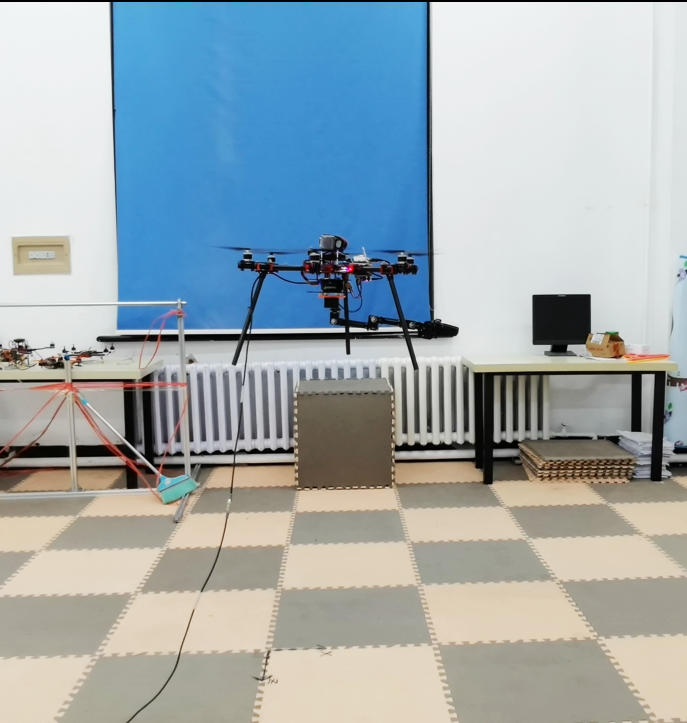}}
		\vspace{2pt}
		\centerline{\includegraphics[width=1.0\textwidth]{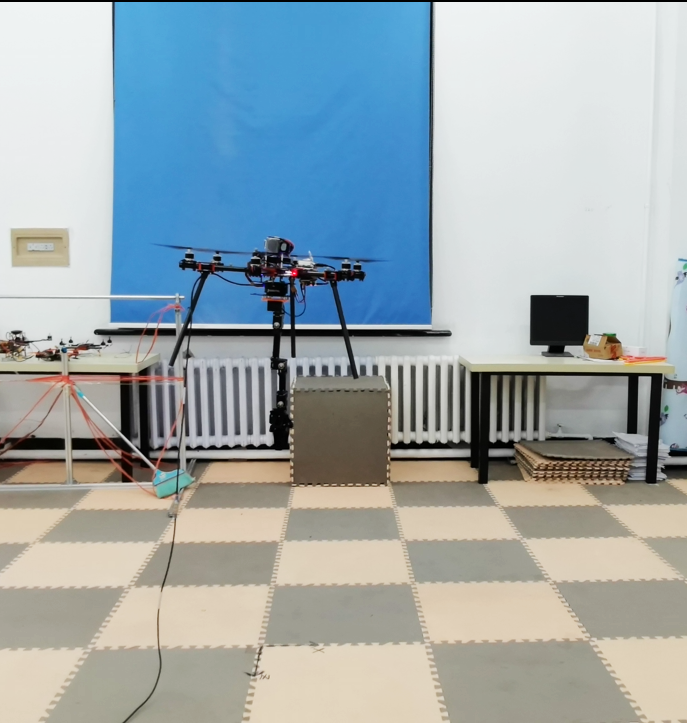}}
	\end{minipage}
	\begin{minipage}{0.16\linewidth}
		\vspace{2pt}
		\centerline{\includegraphics[width=1.0\textwidth]{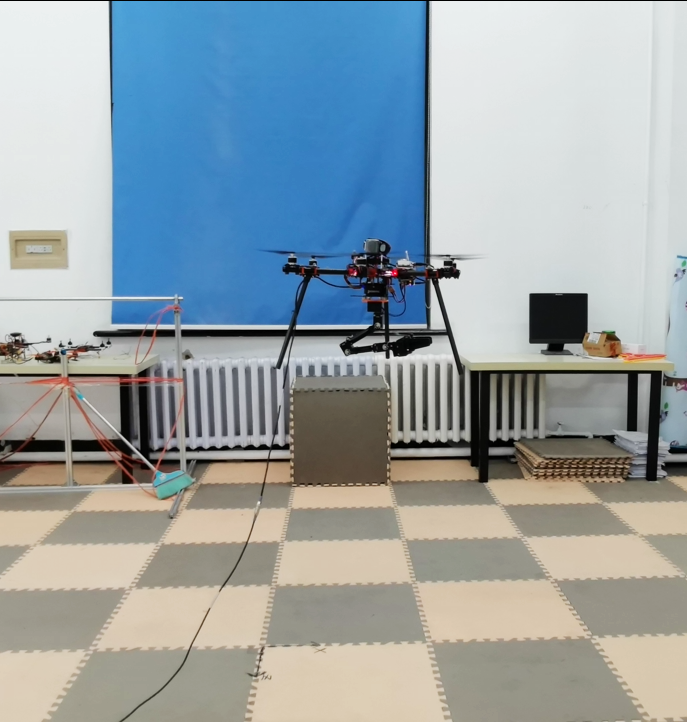}}
		\vspace{2pt}
		\centerline{\includegraphics[width=1.0\textwidth]{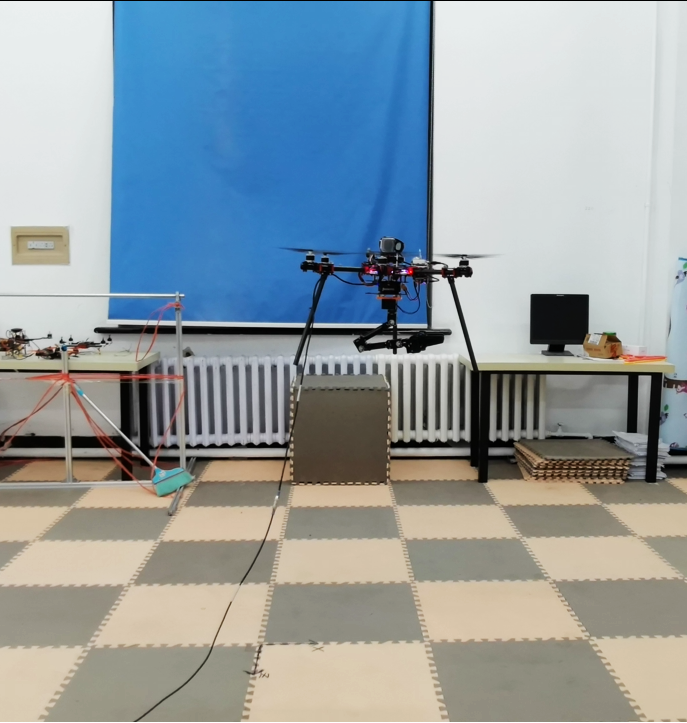}}
	\end{minipage}
	\caption{The snapshots of different stages of the experiment \label{experiment_snapshots}}
\end{figure*}

In order to verify the correctness of the new coupling disturbance model based on variable inertia parameters proposed in Section 2, we carried out physical measurement experiments of the coupling disturbance of AMS. During the experiment, we first let the AMS fly to a height of 1.5m and keep hovering, and then successively let the second joint of the manipulator do sinusoidal swing with the amplitude of $\frac{\pi}{2}$ and the period of 20s and 10s respectively, so that the relative motion between UAV and manipulator is in a large range to let the extreme effects of coupling disturbance can be shown. The motion trajectories and angular velocities of each joint of the manipulator are shown in Fig. \ref{manipulator_motions} as follow. And the snapshots of different stages of the experiment are shown in Fig. \ref{experiment_snapshots}. (The experiment video record can be found in https://youtu.be/vyZVuDXo9Xw)
\subsubsection{Experiment results and analysis}
It is obviously shown from the snapshots of the experiment process in Fig. \ref{experiment_snapshots} that the coupling disturbance force and torque generated when the manipulator moves have a significant impact on the AMS. Especially when the manipulator moves in a large range and at a fast speed, the system attitude changes instantaneously due to the influence of the coupling disturbance torque, resulting in the obvious deviation in the real position of the system from the desired position. The experiment directly proves that the strong coupling disturbance in the AMS will seriously affect the control performance and stability of the system.
\begin{figure}[htb!]
	\centering
	\subfigure[]{\includegraphics[width=0.23\textwidth]{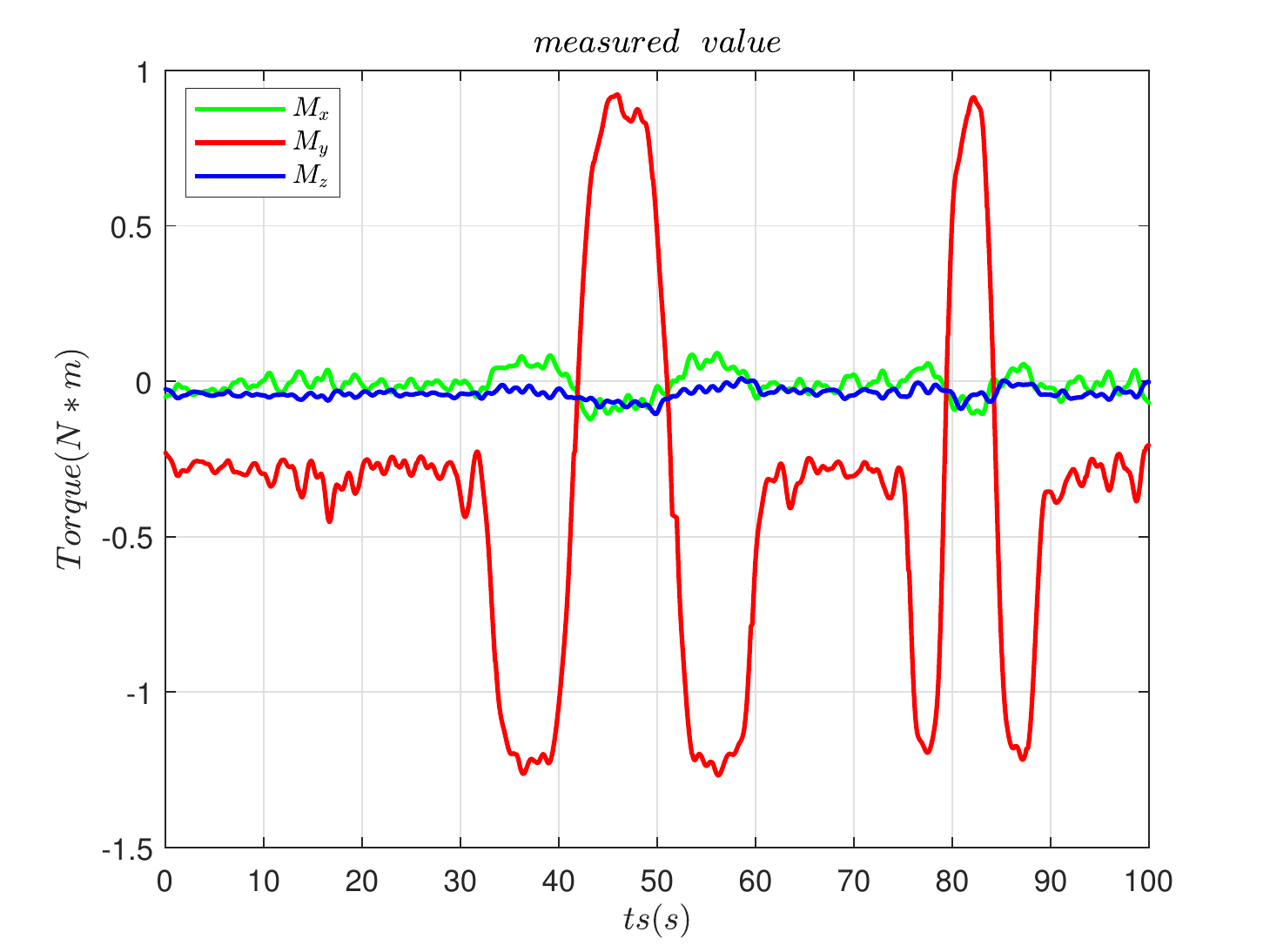}}
	\label{measured_value}
	\subfigure[]{\includegraphics[width=0.23\textwidth]{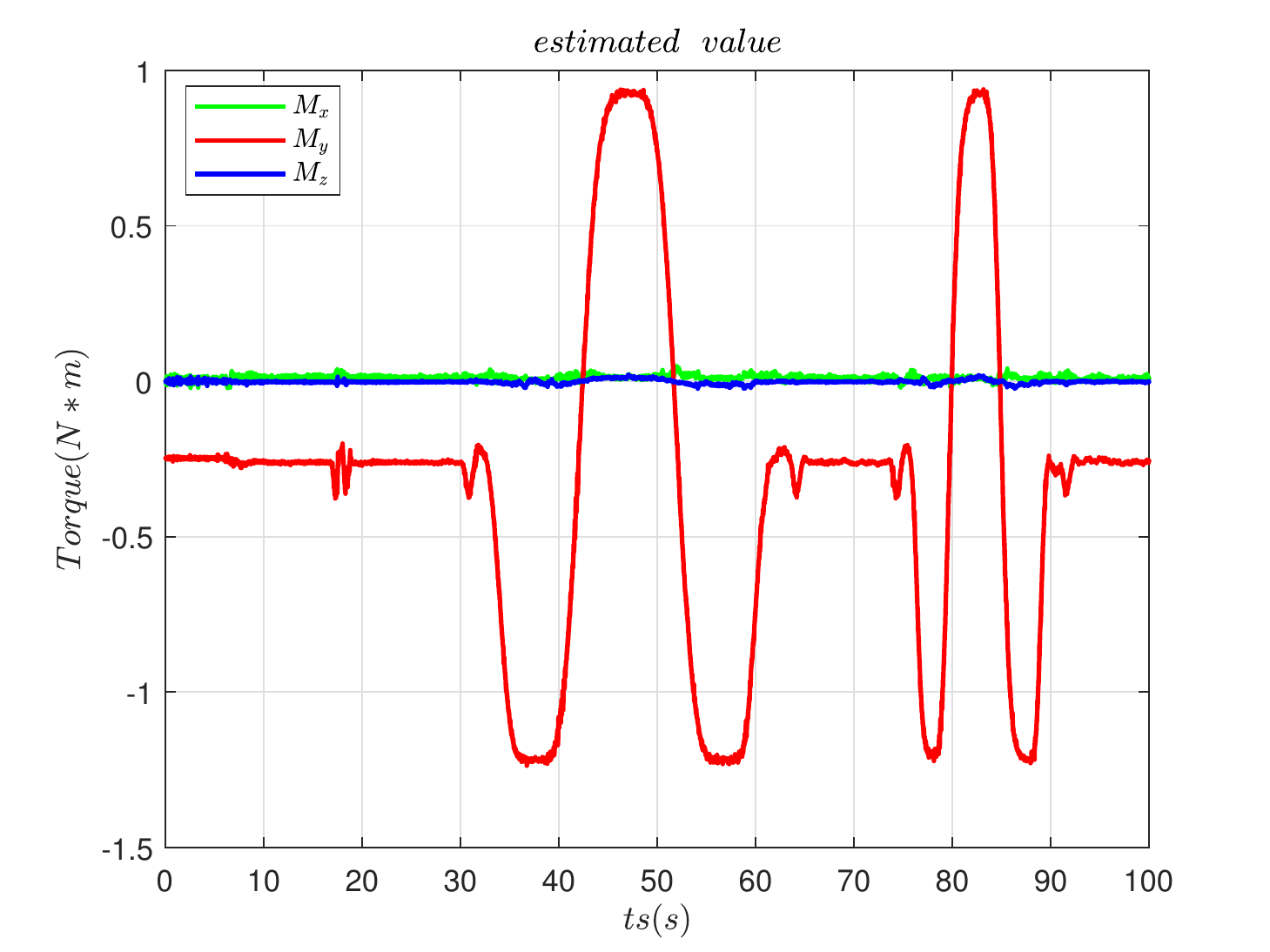}}
	\label{estimated_value}
	\caption{Experimental verification results of coupling disturbance model \label{experiment_result}}
\end{figure}

The experiment results are shown in Fig. \ref{experiment_result}. Fig. \ref{experiment_result}(a)-(b) show the actual coupling disturbance torque measured by the FT300 sensor and the estimated value calculated by the coupling disturbance model, respectively. It can be seen from Fig. \ref{experiment_result} that the estimated value calculated by the coupling disturbance model is very close to the actual measured value. To quantify and compare experimental results, the mean absolute percent error (MAPE) is used to evaluate the error between the model output and measured values, and its index function is defined as follows:
\begin{equation}
MAPE = \frac{1}{N}\sum\limits_{i = 1}^N {\left| {\frac{{{{\hat \tau }_{dis}}\left( i \right) - {\tau _{dis}}\left( i \right)}}{{{\tau _{dis}}\left( i \right)}}} \right|}  \times 100\%
\end{equation}
where, ${{{\hat \tau }_{dis}}\left( i \right)}$ and ${{\tau _{dis}}\left( i \right)}$ represent the estimated value of the coupling disturbance model and actual measured value respectively. ${N}$ denotes the total number of data.

The quantitative comparison results are shown in Table \ref{experiment_error_analysis}. The results show that the coupling disturbance model derived based on variable inertia parameters can effectively and stably estimate the actual coupling disturbance in the AMS with a average residual errors about 8.16\%, which may be caused by some uncertainties or terms that cannot be accurately modeled. Therefore, the coupling disturbance model we derived can contain the dominant part of the actual coupling disturbance in AMS.
\begin{table}[htbp]
	\caption{Experiment Error Analysis}
	\label{experiment_error_analysis}
	\centering
	\begin{tabular}{cc}
		\hline\rule{0pt}{12pt}
		$Parameters$ & $MAPE$ \\
		\hline\rule{0pt}{12pt}
		${}^B{{\hat \tau }_{dis}}$ in X axis & 10.32\% \\
		\hline\rule{0pt}{12pt}
		${}^B{{\hat \tau }_{dis}}$ in Y axis & 5.47\% \\
		\hline\rule{0pt}{12pt}
		${}^B{{\hat \tau }_{dis}}$ in Z axis & 8.69\% \\
		\hline
	\end{tabular}
\end{table}
\subsection{Simulation verification of the proposed control strategy}
\begin{figure*}[b]
	\centering
	\subfigure[]{\includegraphics[width=0.16\textwidth]{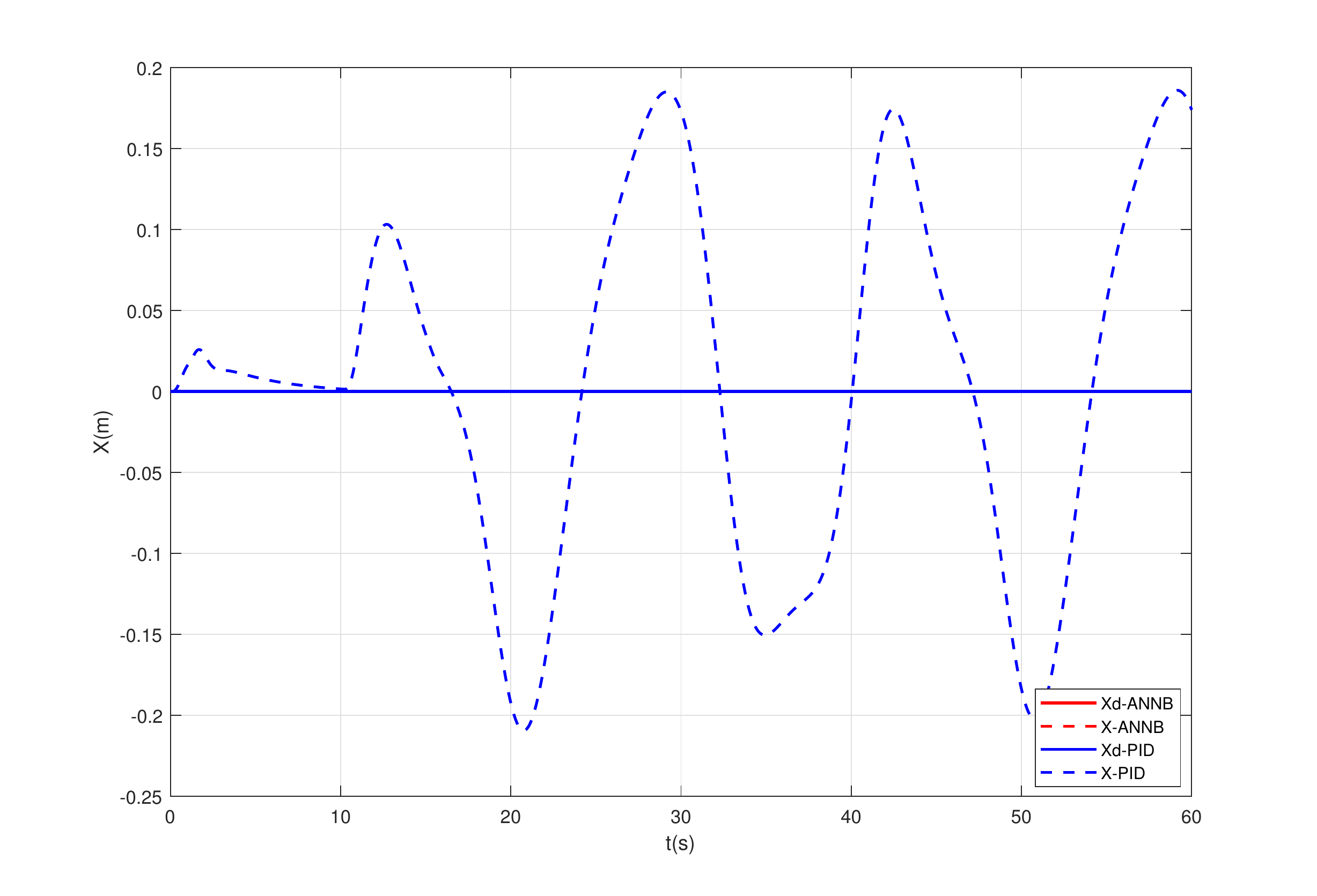}}
	\subfigure[]{\includegraphics[width=0.16\textwidth]{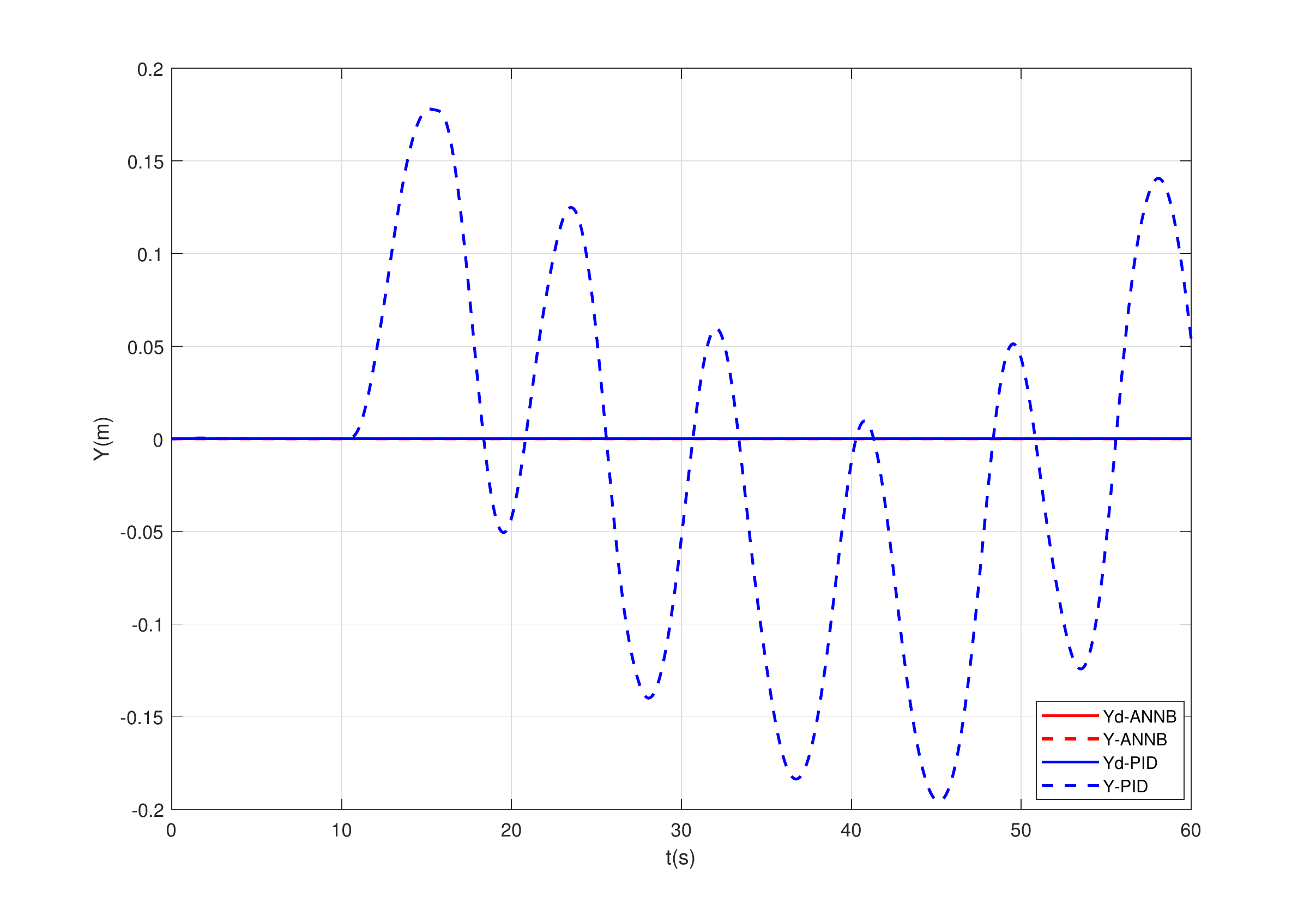}}
	\subfigure[]{\includegraphics[width=0.16\textwidth]{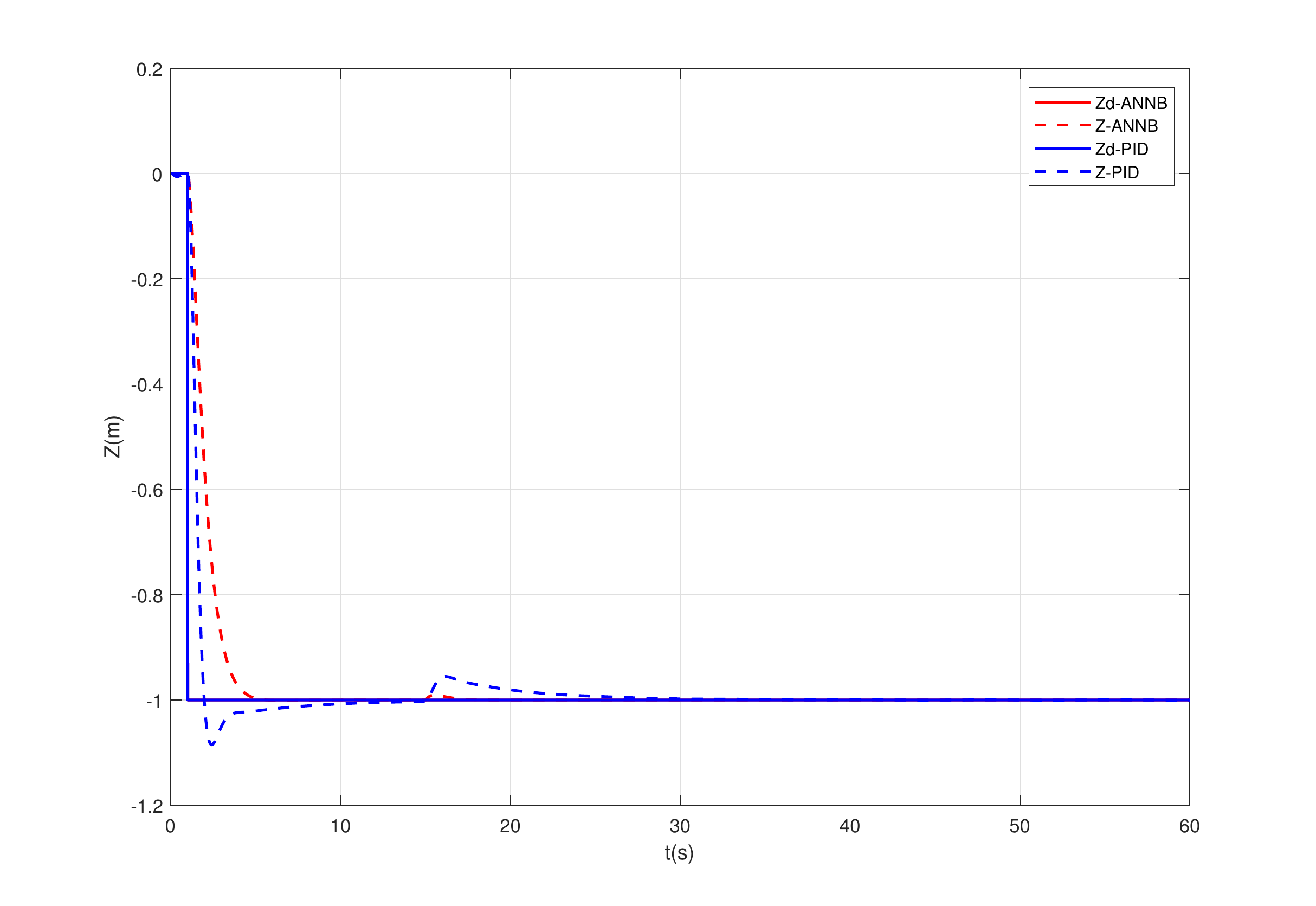}}
	\subfigure[]{\includegraphics[width=0.16\textwidth]{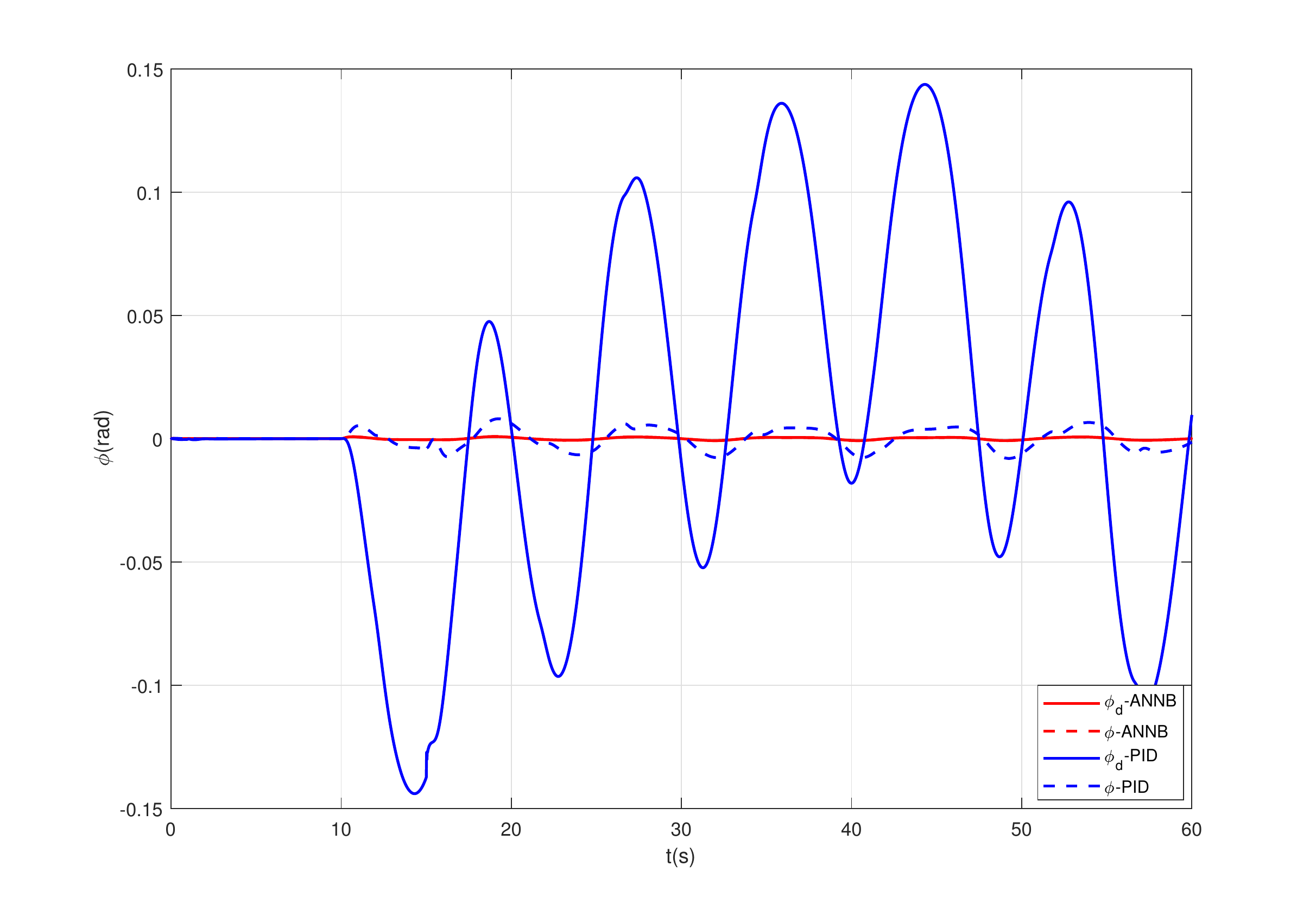}}
	\subfigure[]{\includegraphics[width=0.16\textwidth]{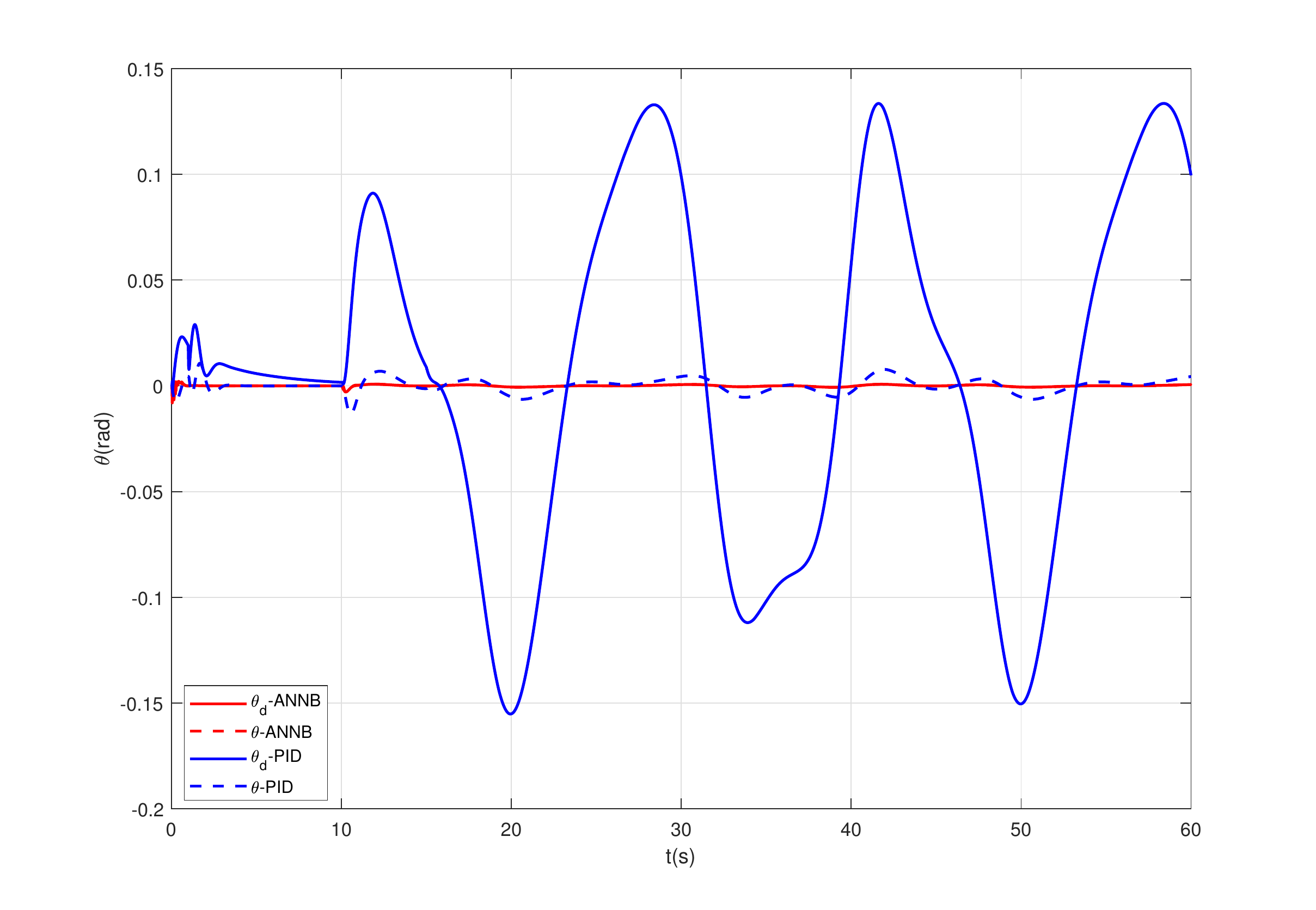}}
	\subfigure[]{\includegraphics[width=0.16\textwidth]{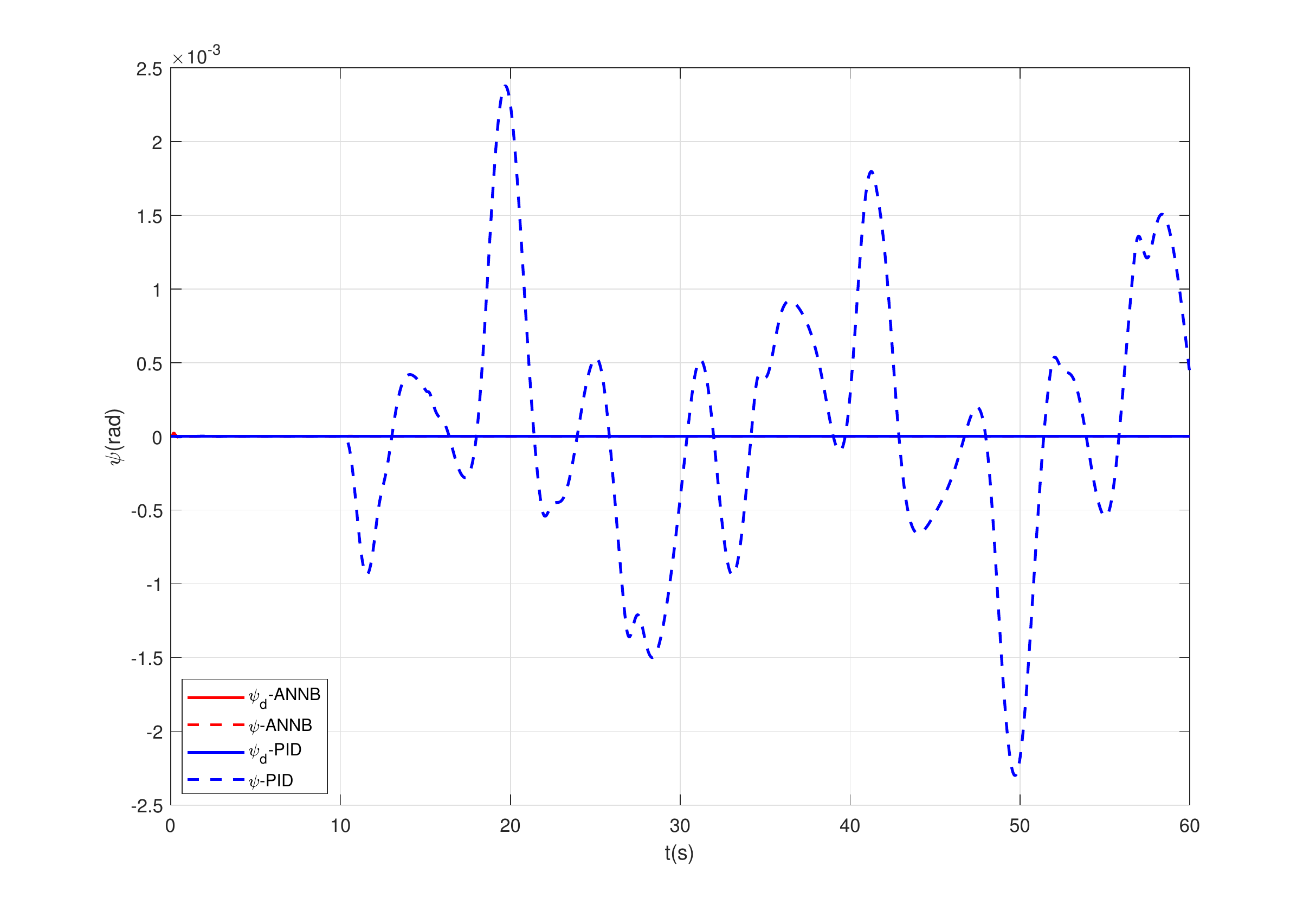}}\\
	\subfigure[]{\includegraphics[width=0.16\textwidth]{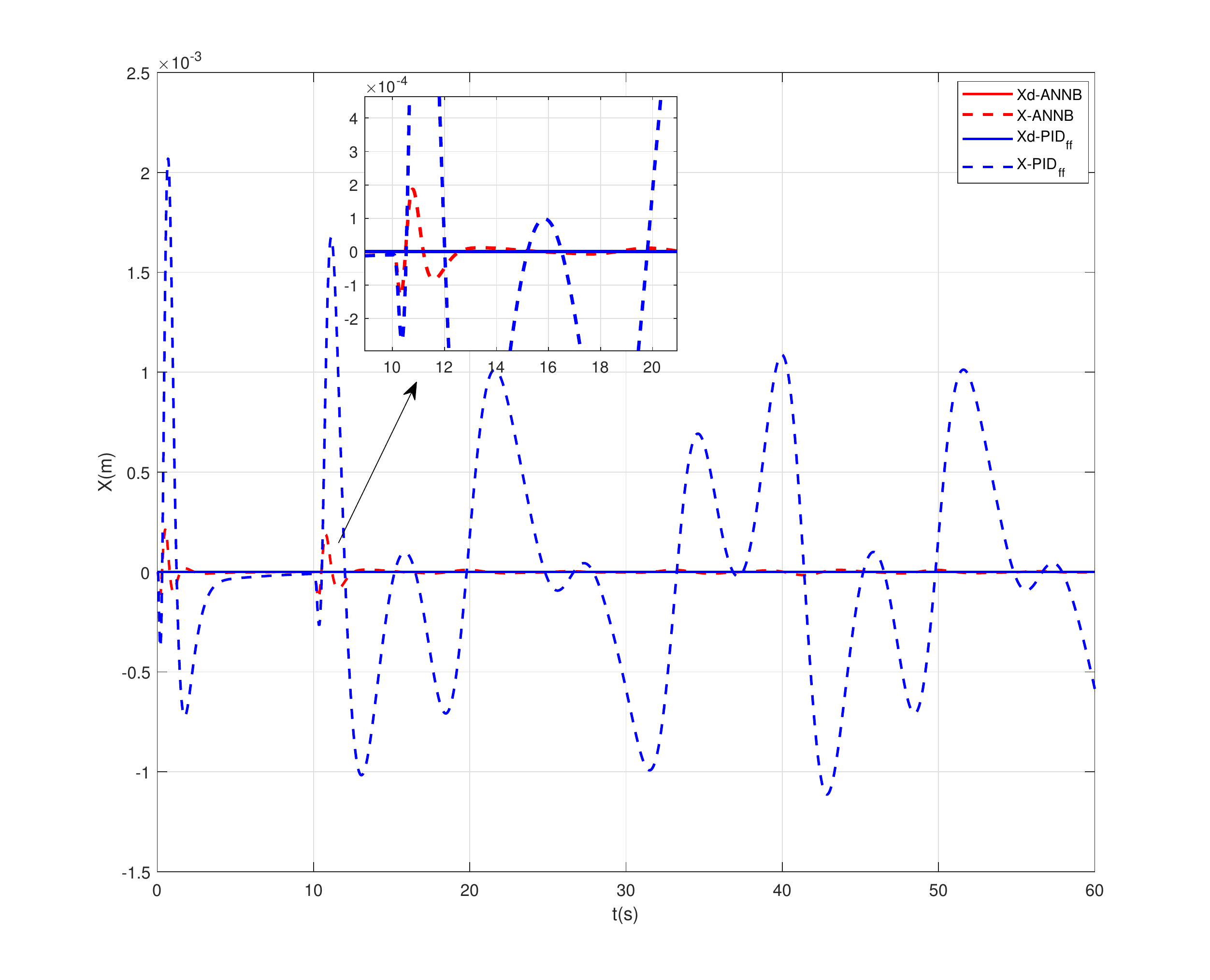}}
	\subfigure[]{\includegraphics[width=0.16\textwidth]{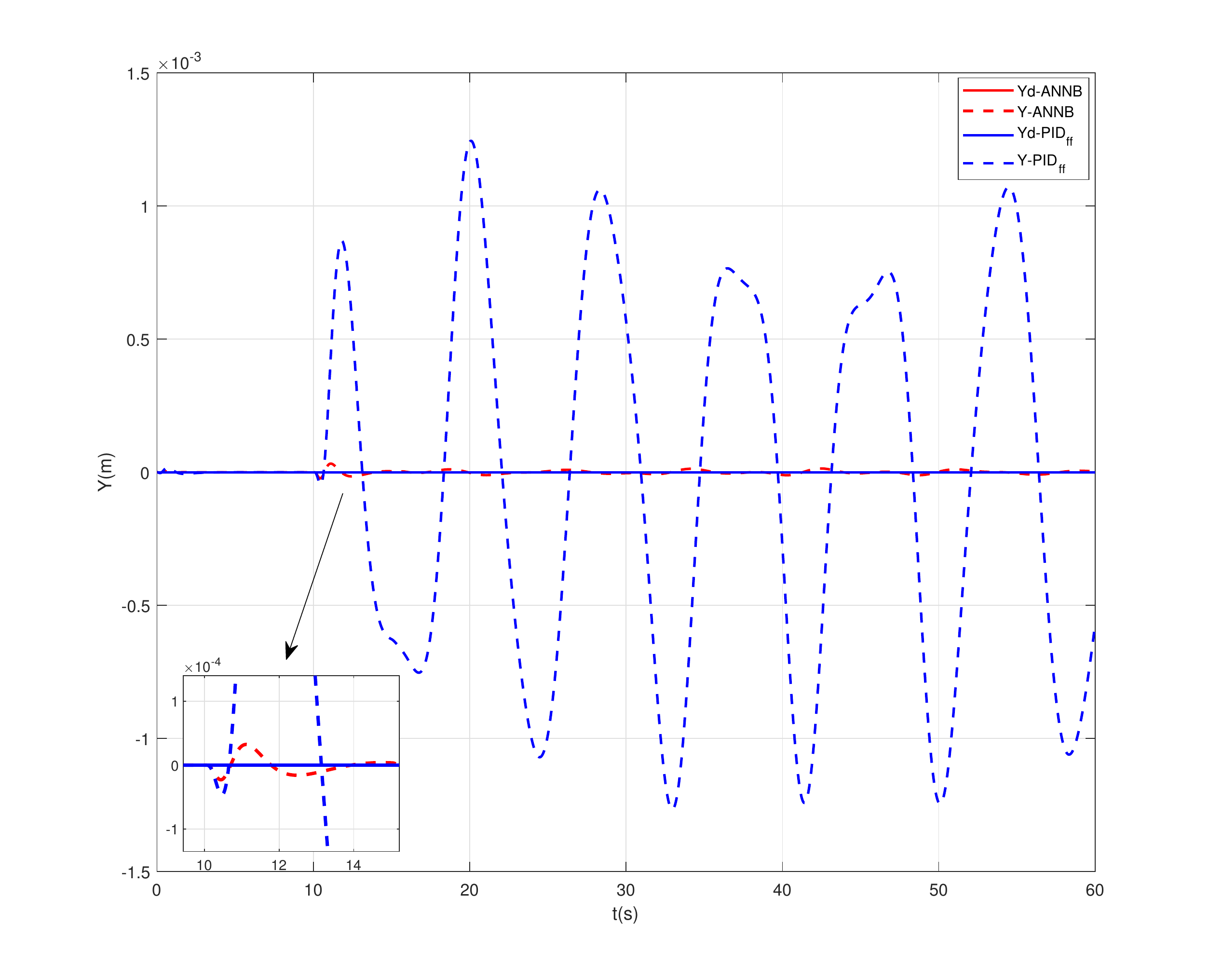}}
	\subfigure[]{\includegraphics[width=0.16\textwidth]{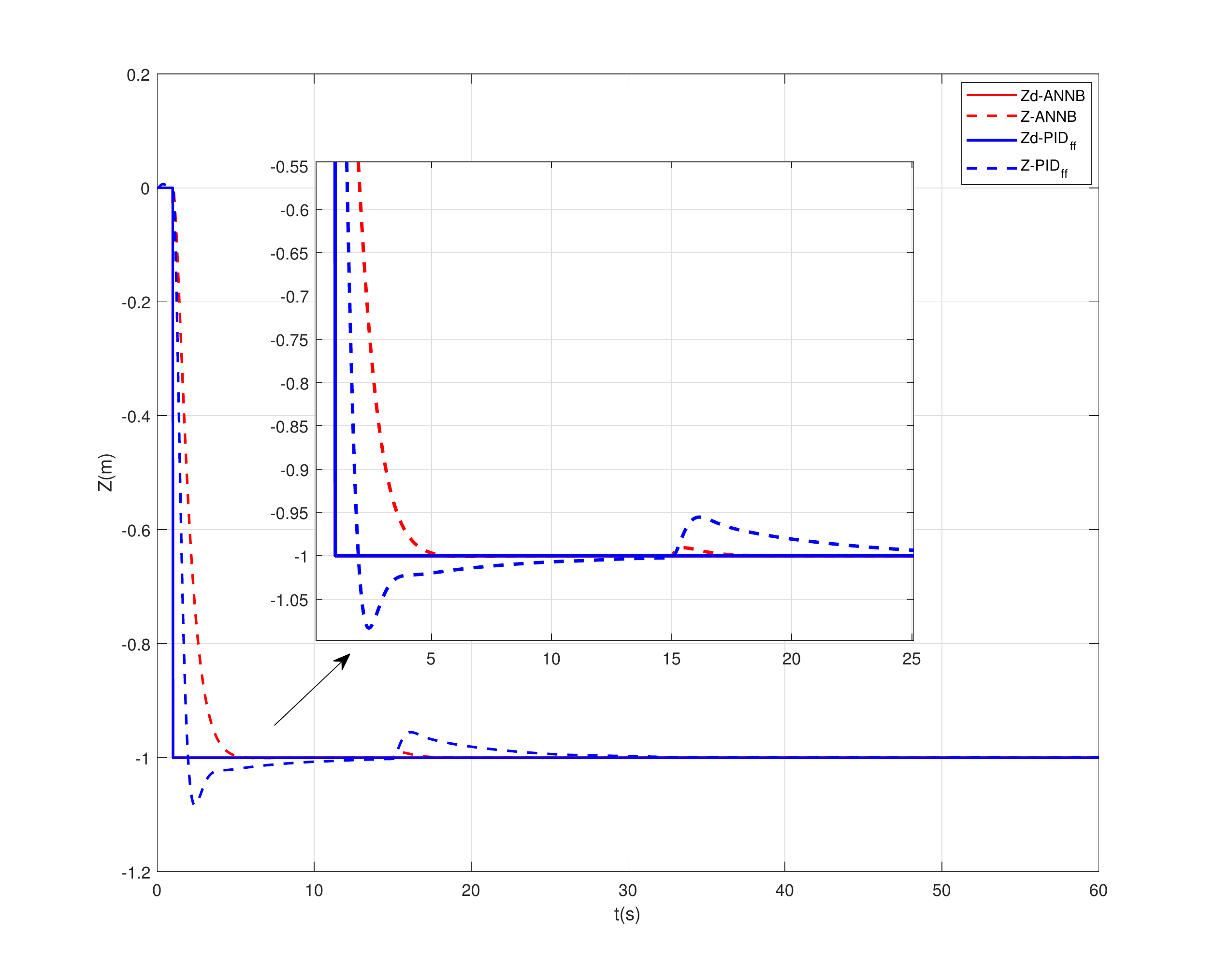}}
	\subfigure[]{\includegraphics[width=0.16\textwidth]{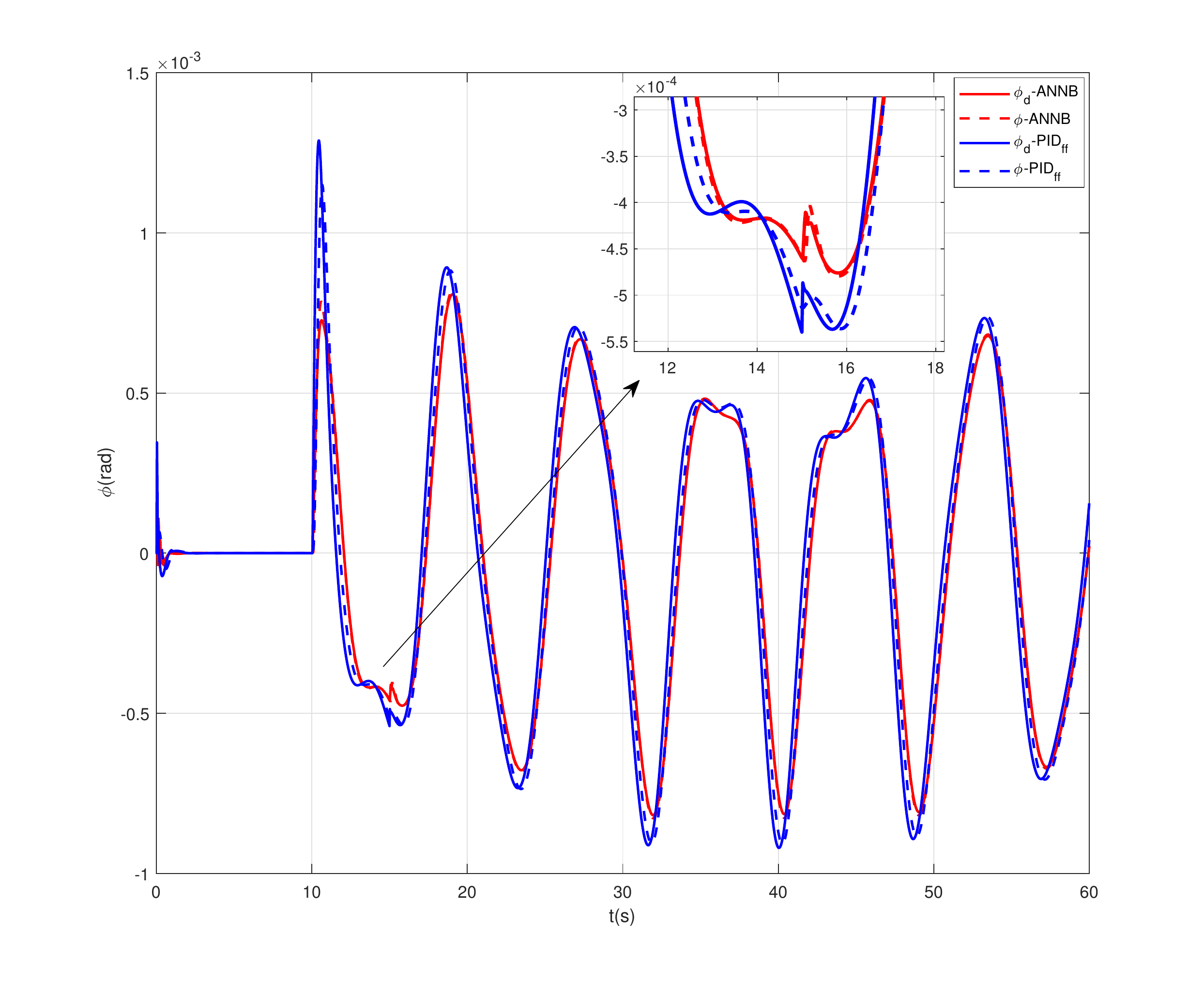}}
	\subfigure[]{\includegraphics[width=0.16\textwidth]{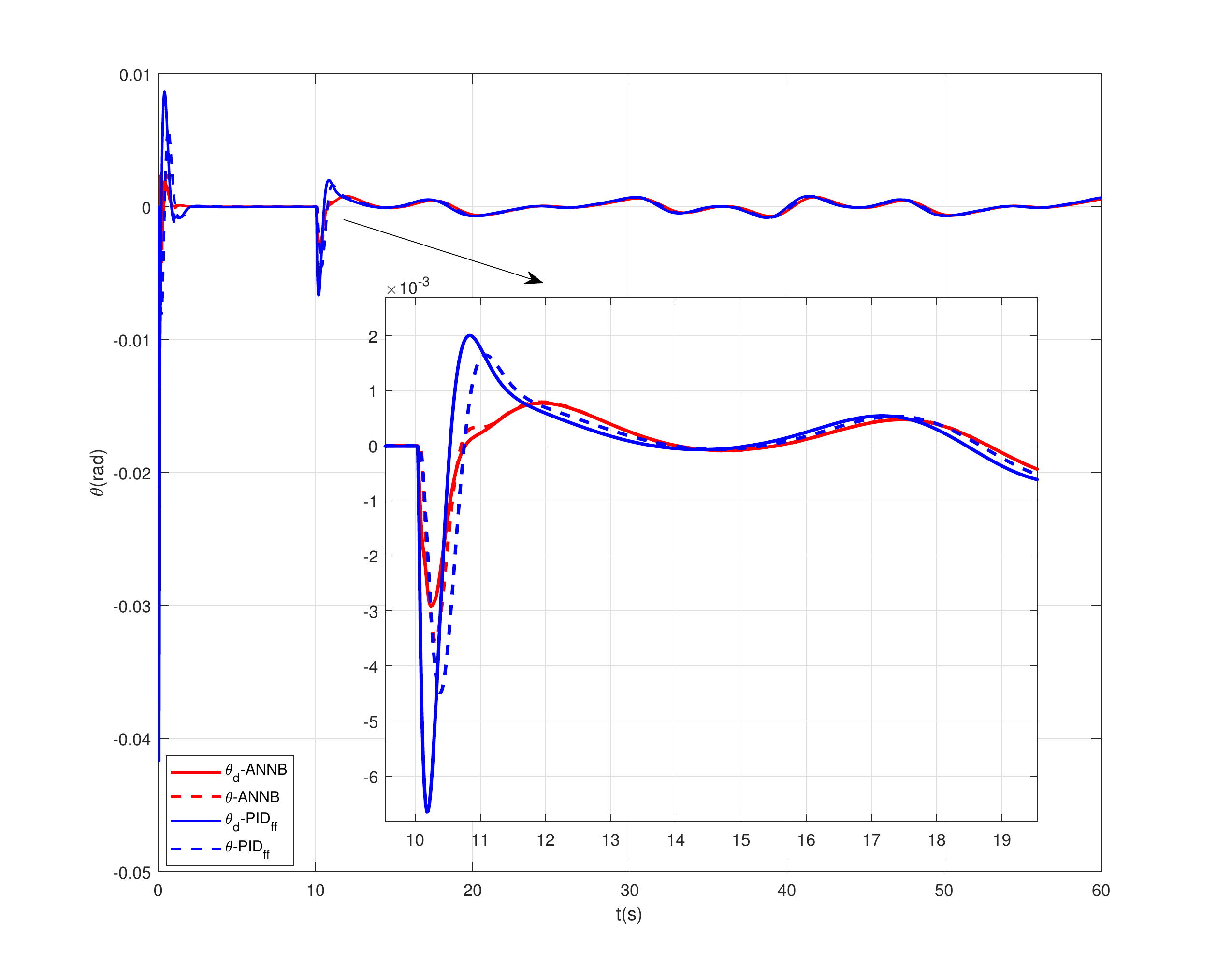}}
	\subfigure[]{\includegraphics[width=0.16\textwidth]{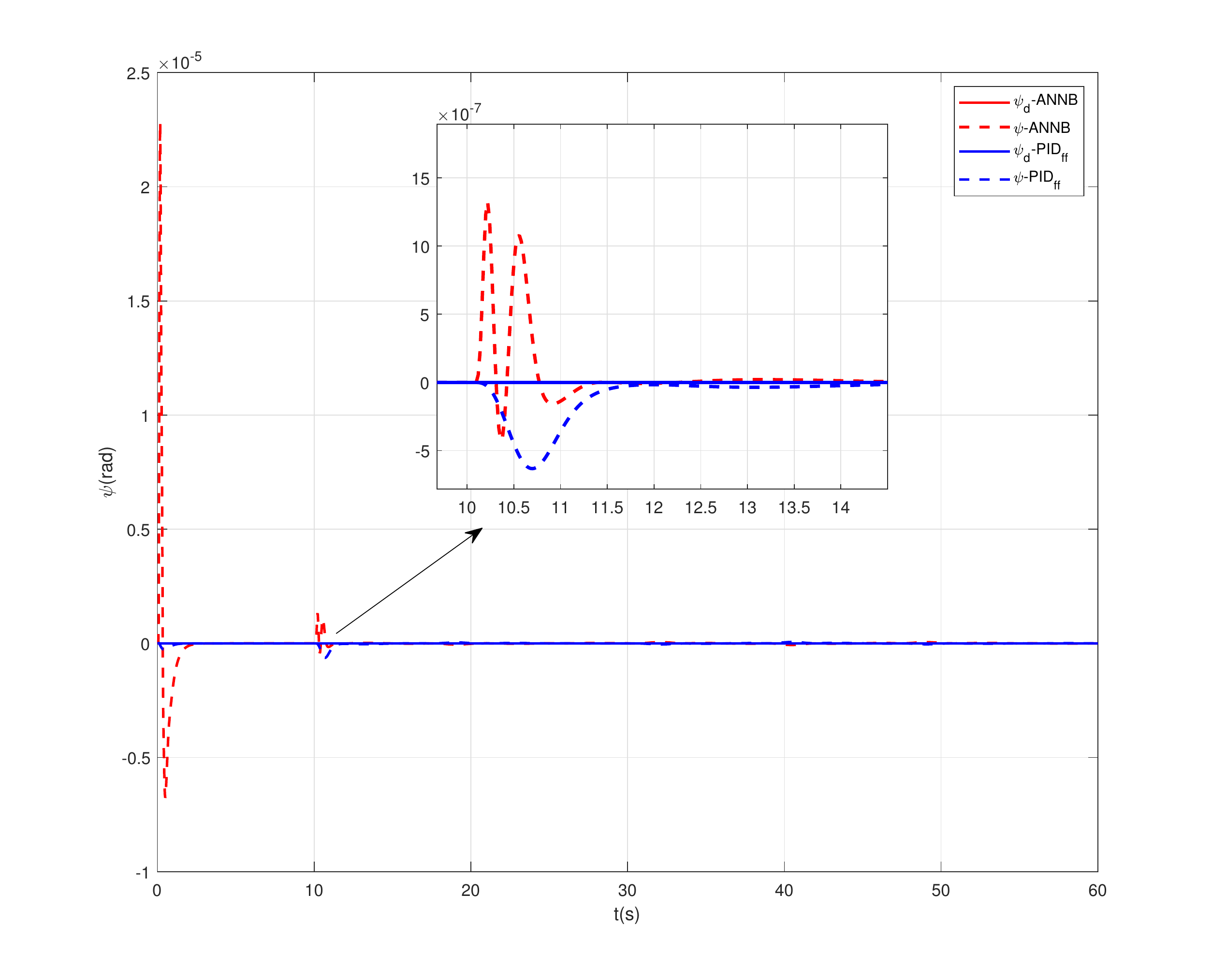}}\\
	\caption{The position and attitude response of the AMS under two sets of comparative simulations ((a) - (f) are for ANNB and PID, (g) - (l) are for ANNB and PID$_{ff}$. solid line is for the desired value, dashed line is for the experiment value) \label{simulation_result_1}}
\end{figure*}
\subsubsection{Simulation conditions}
During the simulation, the initial position of the system is at the origin point of $\Sigma_{I}$, and at 1s, the AMS takes off to a height of 1m and keeps hovering. Then, after the system hovering steadily, at the 10s, let the joint angle of the manipulator operate as the following sinusoidal motion in (\ref{manipulator_motion}). Meanwhile, in order to test the performance of the adaptive neural network, a step additional disturbance with the size of 3.75N is added in the $Z$-axis direction at 15s, which can effectively simulate the additional disturbance caused by manipulator to the AMS when suddenly grasping an object weighing about 375g.
\begin{equation}
\begin{aligned}
{q_1} &= \left\{ {\begin{array}{*{20}{c}}
	{{\rm{             }}0{\rm{             }},t < 10}\\
	{\frac{\pi }{3}\sin \left( {\frac{\pi }{{10}}\left( {t - 10} \right)} \right),t \ge 10}
	\end{array}} \right.,{q_3} =  - \frac{\pi }{2}\\
{q_2} &= \left\{ {\begin{array}{*{20}{c}}
	{{\rm{              }}0{\rm{             }},t < 10}\\
	{\frac{\pi }{3}\sin \left( {\frac{{2\pi }}{{15}}\left( {t - 10} \right)} \right),t \ge 10}
	\end{array}} \right.,{q_4} = 0
\end{aligned}
\label{manipulator_motion}
\end{equation}
The physical parameters of each part of the platform in the simulation are consistent with the parameters of the experiment in the previous section. In addition, the controller parameters used in the simulation are shown in Table \ref{controllers_parameter}.
\begin{table}[htbp]
	\caption{Parameters of Controllers}
	\label{controllers_parameter}
	\centering
	\begin{tabular}{lccc}
		\hline
		\rule{0pt}{12pt}
		PID       & Value    & Our Method    & Value\\[2pt]
		\hline\rule{0pt}{12pt}
		$kp_{xy}$         & 5.0     &$k_{1}$         & 2.0 \\
		\rule{0pt}{0pt}
		$kp_z$            & 4.0       &$k_{2}$         & 0.3\\
		\rule{0pt}{0pt}
		$kp_{v_xv_y}$      & 1.5        &$k_{3}$         & 2.0\\ 
		\rule{0pt}{0pt}
		$kp_{v_z}$       & 5.0        &$k_{4}$         & 0.3\\
		\rule{0pt}{0pt}
		$ki_{v_xv_y}$      & 0.02     &$k_{5}$         & 2.5\\
		\rule{0pt}{0pt}
		$ki_{v_z}$       & 0.02        &$k_{6}$         & 0.9 \\ 
		\rule{0pt}{0pt}
		$kd_{v_xv_y}$      & 1.0        &$k_{7}$         & 4.0 \\ 
		\rule{0pt}{0pt}
		$kd_{v_z}$      & 1.0           &$k_{8}$         & 2.5\\ 
		\rule{0pt}{0pt}
		$kp_{\phi\theta}$         & 6.5    &$k_{9}$         & 4.0 \\
		\rule{0pt}{0pt}
		$kp_{\psi}$            & 3.5      &$k_{10}$         & 2.5\\
		\rule{0pt}{0pt}
		$kp_{pq}$      & 0.8           &$k_{11}$         & 9.2\\ 
		\rule{0pt}{0pt}
		$kp_{r}$       & 2.0    &$k_{12}$         & 3.56\\
		\rule{0pt}{0pt}
		$ki_{pq}$      & 0.2       &$\eta_{1,2}$         & 0.006\\
		\rule{0pt}{0pt}
		$ki_{r}$       & 0.5         &$\eta_{3}$         & 0.04\\ 
		\rule{0pt}{0pt}
		$kd_{pq}$      & 0.03      &$\eta_{4,5}$         & 0.03\\ 
		\rule{0pt}{0pt}
		$kd_{r}$      & 0.02         &$\eta_{6}$         & 0.03 \\[2pt]
		\hline
	\end{tabular}
\end{table}

Based on the above simulation conditions, we compared the anti-disturbance control performance of the proposed method with the PID algorithm and the PID based on the feedforward compensation ofcoupling disturbance model algorithm (PID$_{ff}$) in the two sets of comparative simulations, respectively.

\subsubsection{Simulation results and analysis}
\begin{figure}[htbp]
	\centering
	\subfigure[]{\includegraphics[width=0.15\textwidth]{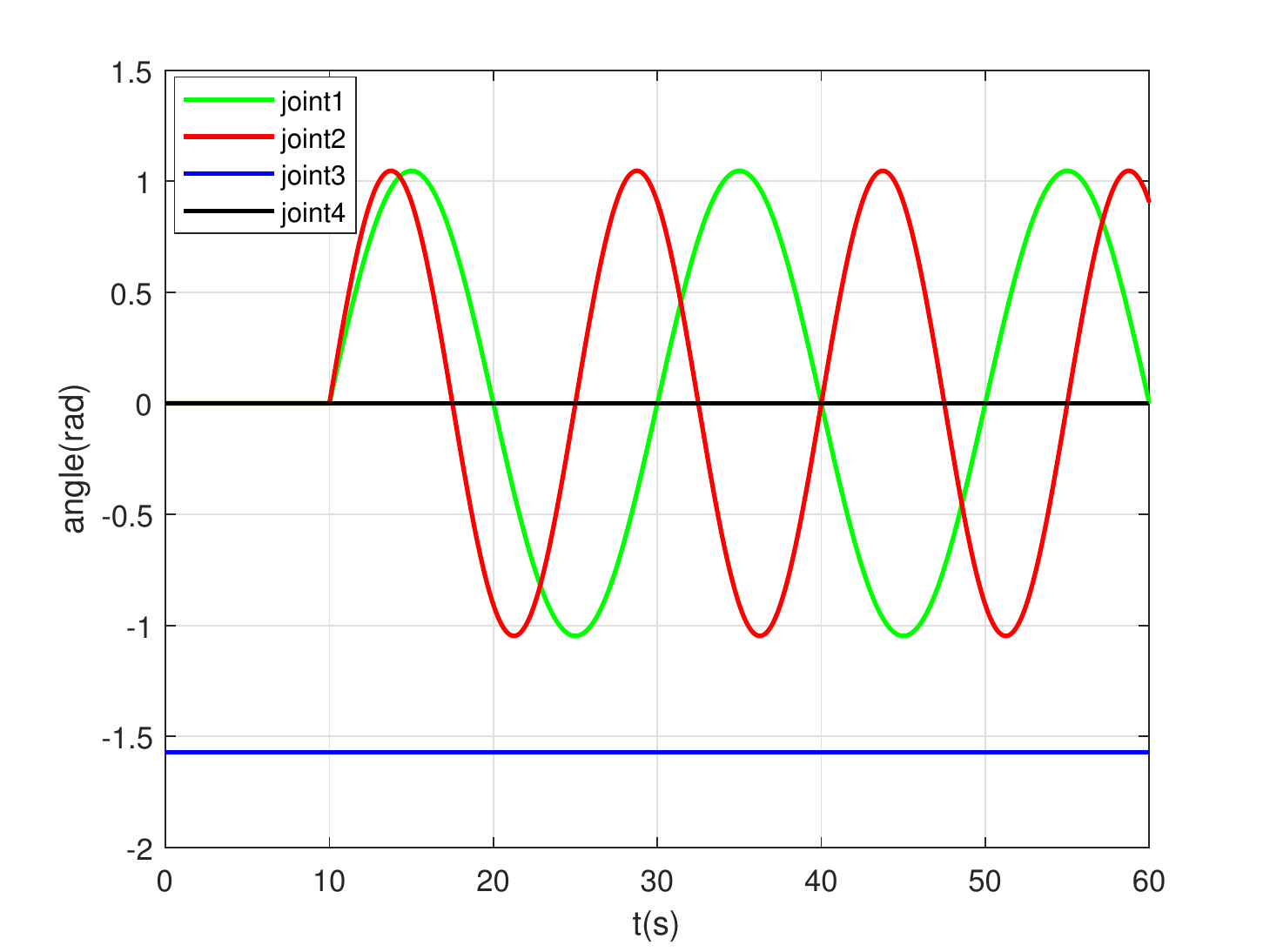}}
	\label{simulation_result_11}
	\subfigure[]{\includegraphics[width=0.15\textwidth]{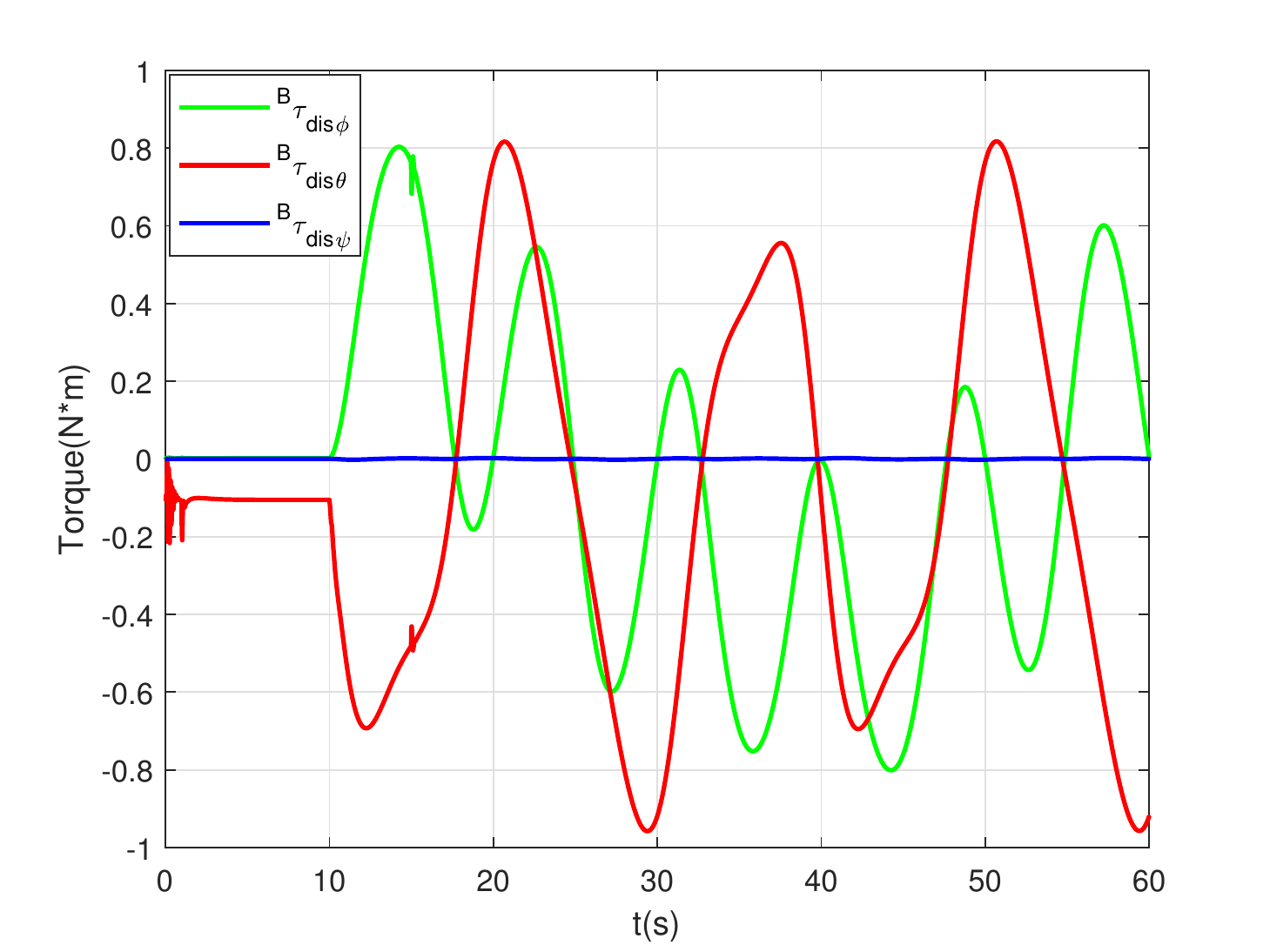}}
	\label{simulation_result_22}
	\subfigure[]{\includegraphics[width=0.15\textwidth]{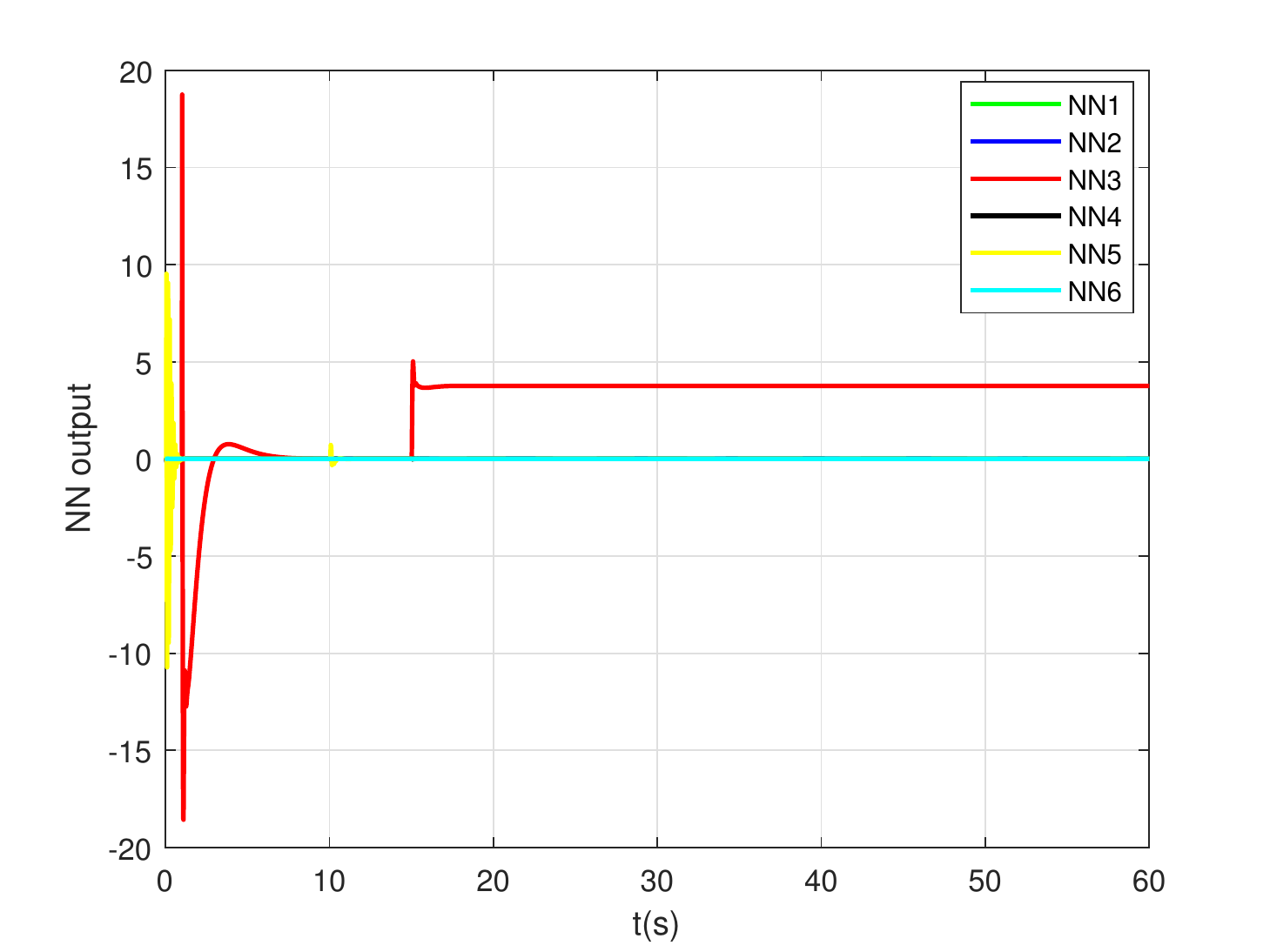}}
	\label{simulation_result_33}\\
	\caption{Simulation results. (a) manipulator joint angles. (b) generated coupling disturbance torque. (c) outputs of adaptive neural networks. \label{simulation_result_2}}
\end{figure}

The simulation results are shown in Fig. \ref{simulation_result_1} and Fig. \ref{simulation_result_2}. Fig. \ref{simulation_result_1}(a)-(c) and Fig. \ref{simulation_result_1}(d)-(f) show the position and attitude response of the AMS under two control methods, our ANNB method and PID algorithm, respectively. Fig. \ref{simulation_result_1}(g)-(i) and Fig. \ref{simulation_result_1}(j)-(l) also show the position and attitude response of the AMS under our ANNB method and PID with feedforward compensation based on the coupled disturbance model algorithm (PID$_{ff}$), respectively. Fig. \ref{simulation_result_2}(a) shows the change of the angle of each joint of the manipulator in the simulations. Fig. \ref{simulation_result_2}(b) shows the coupling disturbance torque estimated by the coupling disturbance model. The neural network output values for the six channels are given in Fig. \ref{simulation_result_2}(c).

From Fig. \ref{simulation_result_1}(a)-(f), we can easily see that the proposed adaptive neural network backstepping control method based on variable inertia parameter modeling has achieved much better performance than traditional cascades PID algorithm for rejecting the strong coupling disturbance in AMS, especially clear from the response of positions after being disturbed. When the manipulator operates in a relatively large range, the influence of the coupling disturbance on the AMS is very obvious, so that it is difficult to resist if only relying on the robustness of the traditional cascade PID algorithm. Due to the strong coupling disturbance torque generated by the large-scale motion of the manipulator, as shown in Fig. \ref{simulation_result_2}(b), the attitude of the AMS changes instantaneously, which is transmitted to the position loop, so that the position of the AMS cannot be maintained in hovering, and a large-scale offset is generated.

By comparing Fig. \ref{simulation_result_1}(a)-(f) and Fig. \ref{simulation_result_1}(g)-(l), it can be found that the cascade PID with feedforward compensation based on the coupled disturbance model (PID$_{ff}$) has significantly improved the anti-disturbance control performance compared with the traditional cascade PID algorithm. Under the influence of strong coupling disturbance, the (PID$_{ff}$) can track the desired signal well in the attitude loop, making the attitude of the AMS stable, thus indirectly ensuring the stability of the position of the AMS, only offset within a few millimeters. It also proves the validity of the coupling disturbance model we proposed. Besides, by comparing the anti-disturbance control performance of the (PID$_{ff}$) with our proposed method, as shown in Fig. \ref{simulation_result_1}(g)-(l), it can be found that our proposed method has achieved relatively better performance, and the specific comparison data can be seen in Table \ref{simulation_error_analysis}.

From Fig. \ref{simulation_result_2}(c), the outputs of adaptive neural network also fully proves that the proposed adaptive neural network can quickly and accurately estimate the additional disturbances caused by unmodeled items or various uncertainties outside the coupling disturbance model. Based on the accurate estimation and compensation of the additional disturbance by the adaptive neural network, our proposed control method has achieved better anti-disturbance control performance compared to PID and (PID$_{ff}$), which can be clearly seen in Fig. \ref{simulation_result_1}(c) and Fig. \ref{simulation_result_1}(i).

To quantitatively compare the simulation effects, the mean error, maximum error, and root mean squared error (RMSE) are used. The quantitative results of the three algorithms are given in Table \ref{simulation_error_analysis}.
\begin{table}[htbp]
	\caption{Simulation Error Analysis}
	\label{simulation_error_analysis}
	\centering
	\begin{tabular}{|l|c|c|c|}
		\hline
		\rule{0pt}{12pt}
		& Mean      & Maximum   & RMSE  \\[2pt] 
		\hline\rule{0pt}{12pt}
		$X$:PID     & 0.088891    & 0.209327    & 0.110359  \\
		\rule{0pt}{0pt}
		$X$:PID$_{ff}$     & 0.000407    & 0.002070    & 0.000559  \\
		\rule{0pt}{0pt}
		$X$:Our Method   & 0.000008  & 0.000212   & 0.000024 \\
		\rule{0pt}{0pt}
		$Y$:PID     & 0.072511    & 0.195585    & 0.094819  \\
		\rule{0pt}{0pt}
		$Y$:PID$_{ff}$     & 0.000547    & 0.001267    & 0.000670  \\
		\rule{0pt}{0pt}
		$Y$:Our Method   & 0.000004  & 0.000032   &  0.000006 \\
		\rule{0pt}{0pt}
		$Z$:PID     & 0.015600    & 1.002806    & 0.077819  \\
		\rule{0pt}{0pt}
		$Z$:PID$_{ff}$     & 0.015247    & 0.996862    & 0.076178  \\
		\rule{0pt}{0pt}
		$Z$:Our Method   & 0.017936  & 1.000000   & 0.104279 \\
		\rule{0pt}{0pt}
		$\phi$:PID     & 0.053729    & 0.140313    & 0.069754  \\
		\rule{0pt}{0pt}
		$\phi$:PID$_{ff}$     & 0.000069    & 0.000682    & 0.000099  \\
		\rule{0pt}{0pt}
		$\phi$:Our Method   & 0.000004  & 0.000130   & 0.000009 \\
		\rule{0pt}{0pt}
		$\theta$:PID     & 0.065027    & 0.150169    & 0.079837  \\
		\rule{0pt}{0pt}
		$\theta$:PID$_{ff}$     & 0.000196    & 0.041667    & 0.001278  \\
		\rule{0pt}{0pt}
		$\theta$:Our Method   & 0.000031  & 0.008098   & 0.000305 \\
		\rule{0pt}{0pt}
		$\psi$:PID     & 0.000561    & 0.002377    & 0.000800  \\
		\rule{0pt}{0pt}
		$\psi$:PID$_{ff}$     & 0.000001    & 0.000001    & 0.000001  \\
		\rule{0pt}{0pt}
		$\psi$:Our Method   & 0.000001  & 0.000022   & 0.000001 \\[2pt]
		\hline
	\end{tabular}
\end{table}

\section{Conclusion}
\label{Conclusion}
In this paper, aiming at the problem of anti-disturbance control of the AMS facing multi-source disturbances when performing aerial work tasks, we propose an anti-disturbance control strategy based on coupling disturbance modeling for feedforward compensation and adaptive neural network estimation for feedback compensation. And based on this, an adaptive neural network backstepping control method based on variable inertia parameter modeling is proposed. First, the coupling disturbance model is derived based on the variable inertia parameters. Through the coupling disturbance model, we can compensate the strong coupling disturbance in a way of feedforward. Then, to estimate and deal with additional disturbances caused by unmodeled dynamic terms and various uncertainties, a feedback compensation method based on adaptive neural network estimation is proposed. Finally, the effectiveness of the proposed method is verified by experiments and simulations. The experimental results show that the proposed coupling disturbance model can accurately obtain the coupling disturbance in the AMS. The simulation results also show that the proposed control strategy can effectively solve the problem of anti-multi-source disturbances in the AMS, and obtain better performance than the PID algorithm and PID algorithm with coupling disturbance feedforward compensation (PID$_{ff}$).

In the future work, we will carry out experiments on the physical platform to verify the feasibility and effectiveness of the proposed disturbance rejection control strategy and try to find ways to reduce the coupling disturbance effection on the system as much as possible from the perspective of motion planning method.

\bibliographystyle{IEEEtran}
\bibliography{editor}

%
%
%
%
\vspace{-25 pt}
\begin{IEEEbiography}[{\includegraphics[width=1in,height=1.25in,clip,keepaspectratio]{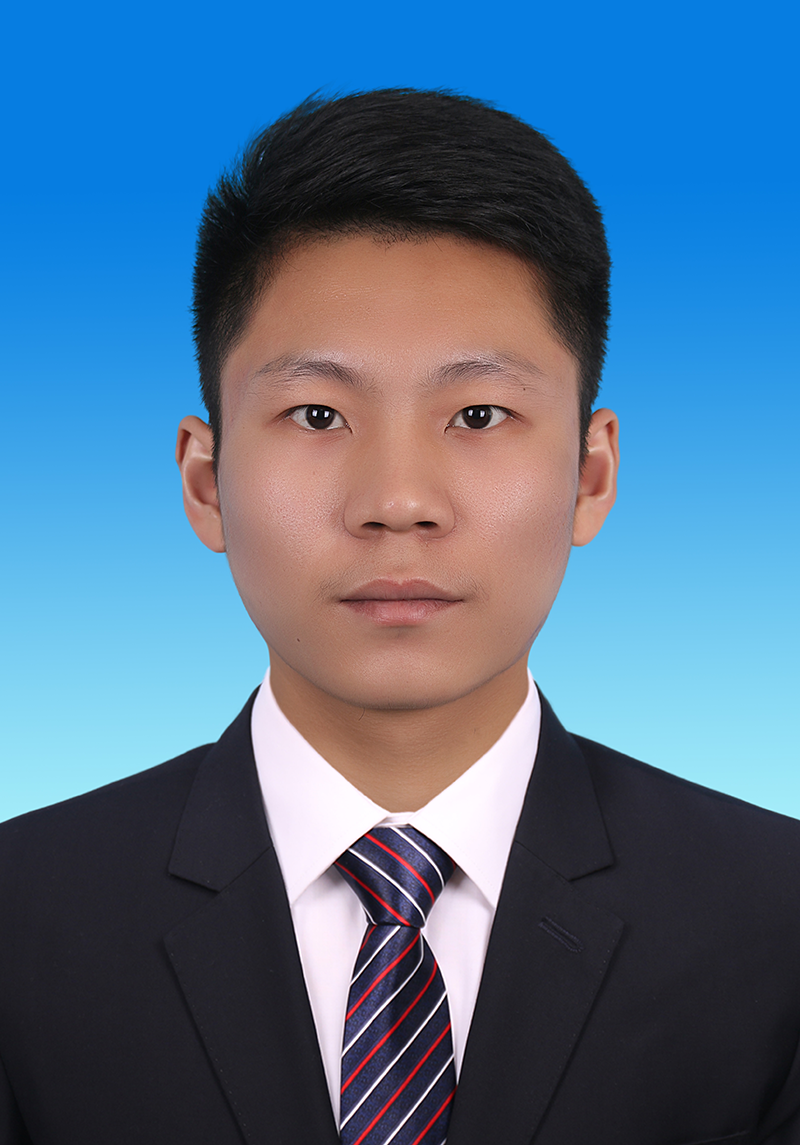}}]{Hai Li}
	was born in Sichuan, China, in 1996. He received the B.S. degree in aircraft design and engineering from the Harbin Engineering University, Harbin, China, in 2018. He is currently working toward the Ph.D. degree in the School of Astronautics, Harbin Institute of Technology, Harbin, China.
	
	His research interests include dynamic modeling, control and planning of the aerial manipulator.
\end{IEEEbiography}
\vspace{-25 pt}
\begin{IEEEbiography}[{\includegraphics[width=1in,height=1.25in,clip,keepaspectratio]{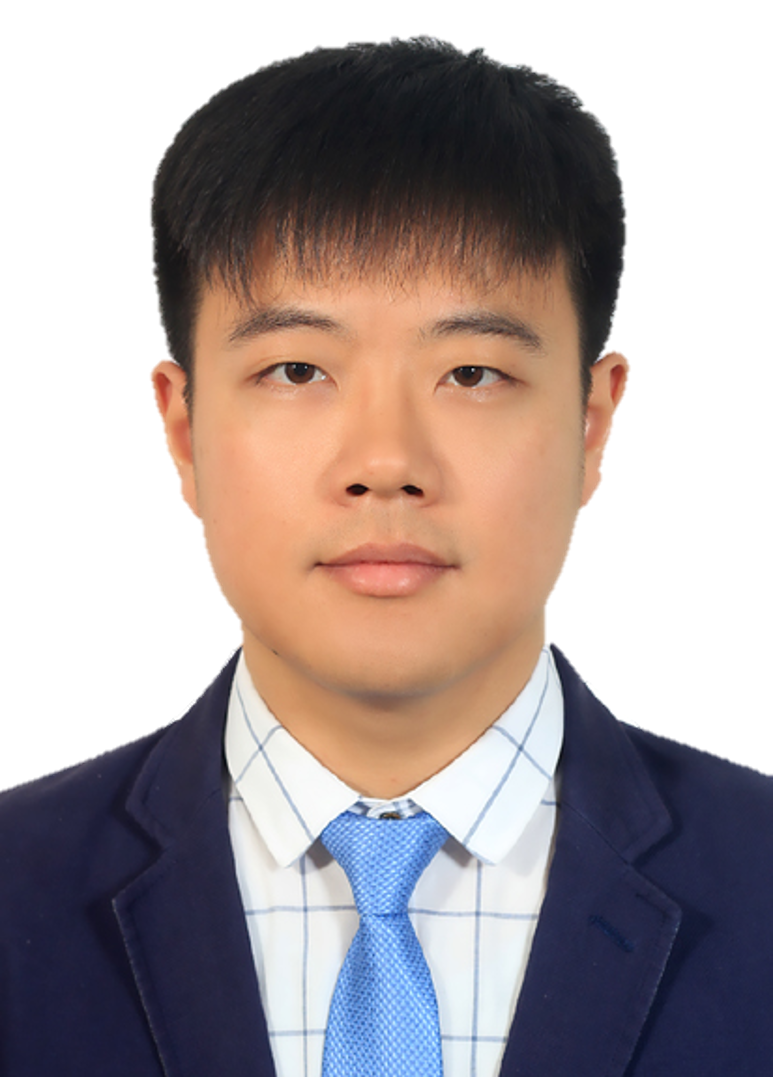}}]{Zhan Li}
	(M'16) received the Ph.D. degree in control science and engineering from the Harbin Institute of Technology, Harbin, China, in 2015.
	
	He is currently an Associate Professor with the Research Institute of Intelligent Control and Systems, School of Astronautics, Harbin Institute of Technology. His research interests include motion control, industrial robot control, robust control of small unmanned aerial vehicles (UAVs), and cooperative control of multivehicle systems.
\end{IEEEbiography}
\vspace{-25 pt}
\begin{IEEEbiography}[{\includegraphics[width=1in,height=1.25in,clip,keepaspectratio]{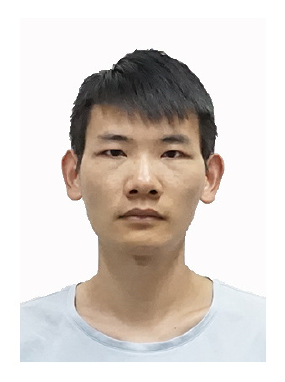}}]{Xiaolong Zheng}
	received the B.S. degree in automation from Yangtze University College of Technology and Engineering, Jingzhou, China, in 2013, the M.S. degree in control theory and control engineering from Bohai University, Jinzhou, China, in 2016, and the Ph.D. degree in control science and engineering from Harbin Institute of Technology, Harbin, China, in 2020.
	
	He is currently an Assistant Professor with the Research Institute of Intelligent Control Systems, Harbin Institute of Technology. His research interests include adaptive control, neural networks, reinforcement learning and their applications.
\end{IEEEbiography}
\vspace{-25 pt}
\begin{IEEEbiography}[{\includegraphics[width=1in,height=1.25in,clip,keepaspectratio]{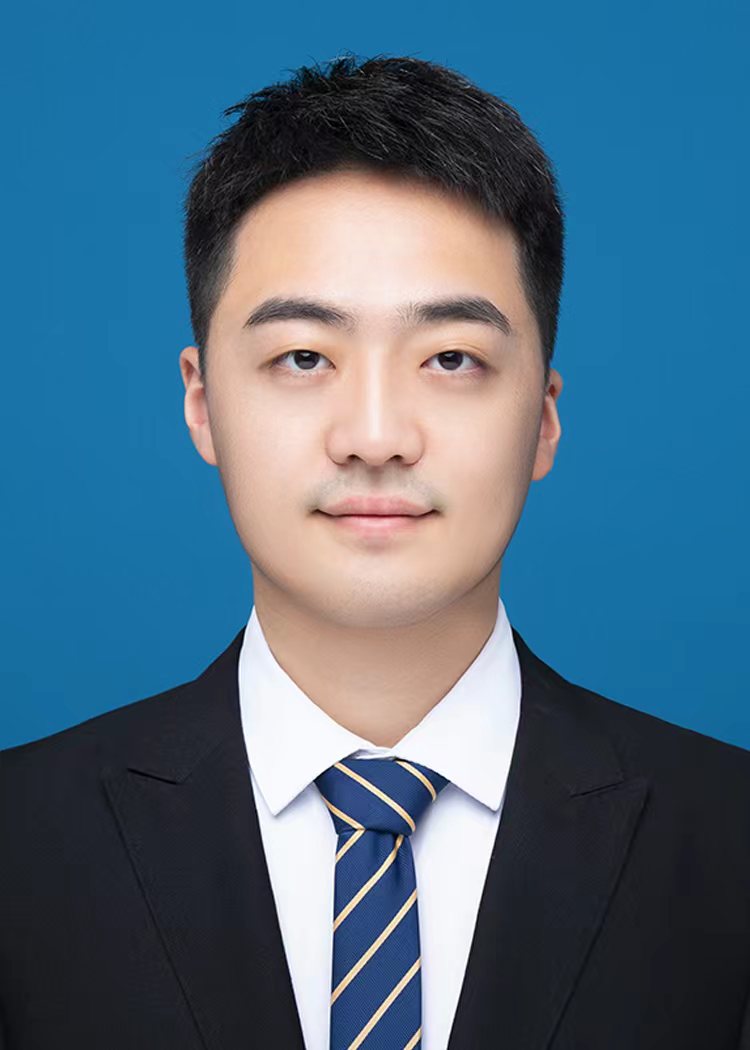}}]{Jinhui Liu}	
	was born in Jiangsu, China, on May 8, 1997. He received the B.S. degree in control science and technology from Nanjing University of Science and Technology, Nanjing, China, in 2019. He received the M.S. degree in control science and technology from Harbin Institute of Technology, Harbin, China, in 2021. He is currently working for the Ph.D. degree in the School of Aeronautics, Harbin Institute of Technology,  Harbin, China.
	
	His research interests include control and planning of aerial manipulator system and multi-agent system.
\end{IEEEbiography}

\end{document}